\documentclass{article}
\pdfoutput=1

\usepackage{arxiv}

\usepackage[utf8]{inputenc} 
\usepackage[T1]{fontenc}    
\usepackage{hyperref}       
\usepackage{url} 
\usepackage{booktabs}       
\usepackage{amsfonts}       
\usepackage{nicefrac}       
\usepackage{microtype}      
\usepackage{lipsum}
\usepackage{graphicx}
\usepackage{listings}
\usepackage{textcomp}
\usepackage{tabularx}
\usepackage[flushleft]{threeparttable}
\usepackage{multirow}
\usepackage{multicol}
\usepackage{array}
\usepackage{float}
\usepackage{xcolor}
\usepackage{amsmath}
\usepackage{pgfplots}
\usetikzlibrary{patterns}

\hypersetup{
  colorlinks = true,
  citecolor = blue,
  urlcolor = blue
}

\newcommand{\etal}[0]{\textit{et al.}}
\newcommand{\ie}[0]{i.e.}
\newcommand{\eg}[0]{e.g.}

\title{Towards the Generation of Synthetic Images of Palm Vein Patterns: A Review}

\author{
 Edwin H. Salazar-Jurado\href{https://orcid.org/0000-0002-0731-0779}{\hspace{1mm}\includegraphics[scale=0.09]{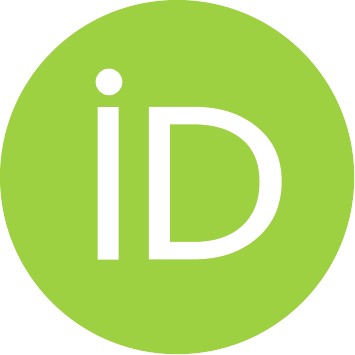}} \\
  Doctorado en Modelamiento Matemático Aplicado\\
  Universidad Católica del Maule\\
  Talca 3480112, Chile \\
  \texttt{edwin.salazar@alu.ucm.cl} \\
   \And
  Ruber Hern\'andez-Garc\'ia\href{https://orcid.org/0000-0002-9311-1193}{\hspace{1mm}\includegraphics[scale=0.09]{orcid.pdf}}\\
  Laboratory of Technological Research in Pattern Recognition\\
  Research Center for Advanced Studies of Maule\\
  Universidad Católica del Maule\\
  Talca 3480112, Chile \\
  \texttt{rhernandez@ucm.cl} \\
  \And
  Karina Vilches-Ponce\href{https://orcid.org/0000-0001-7689-1696}{\hspace{1mm}\includegraphics[scale=0.09]{orcid.pdf}} \\
  Faculty of Basic Sciences\\
  Universidad Católica del Maule\\
  Talca 3480112, Chile \\
  \texttt{kvilches@ucm.cl} \\
  \And
  Ricardo J. Barrientos\href{https://orcid.org/0000-0001-5345-7061}{\hspace{1mm}\includegraphics[scale=0.09]{orcid.pdf}}\\
  Laboratory of Technological Research in Pattern Recognition\\
  Faculty of Engineering Sciences\\
  Universidad Católica del Maule\\
  Talca 3480112, Chile \\
  \texttt{rbarrientos@ucm.cl} \\
  \And
  Marco Mora\href{https://orcid.org/0000-0003-3619-2561}{\hspace{1mm}\includegraphics[scale=0.09]{orcid.pdf}}\\
  Laboratory of Technological Research in Pattern Recognition\\
  Faculty of Engineering Sciences\\
  Universidad Católica del Maule\\
  Talca 3480112, Chile \\
  \texttt{mmora@ucm.cl} \\
  \And
  Gaurav Jaswal\href{https://orcid.org/0000-0002-3971-0160}{\hspace{1mm}\includegraphics[scale=0.09]{orcid.pdf}}\\
  Department of Electrical Engineering\\
  Indian Institute of Technology Delhi\\
  New Delhi 110016, India \\
  \texttt{gauravjaswal@iitd.ac.in} \\
}

\date{}

\renewcommand{\headeright}{Preprint Under Review}

\hypersetup{
pdftitle={Towards the Generation of Synthetic Images of Palm Vein Patterns: A Review},
pdfsubject={biometrics, computer vision, mathematical modeling},
pdfauthor={Edwin H. Salazar-Jurado, Ruber Hern\'andez-Garc\'ia, Karina Vilches-Ponce, Ricardo J. Barrientos, Marco Mora, Gaurav Jaswal},
pdfkeywords={Biometrics, Image synthesis, Palm vein recognition, Palm vein datasets, Synthetic palm vein images},
}

\begin{document}
\maketitle
\thispagestyle{firstpage}
\begin{abstract}
With the recent success of computer vision and deep learning, remarkable progress has been achieved on automatic personal recognition using vein biometrics. However, collecting large-scale real-world training data for palm vein recognition has turned out to be challenging, mainly due to the noise and irregular variations included at the time of acquisition. Meanwhile, existing palm vein recognition datasets are usually collected under near-infrared light, lacking detailed annotations on attributes (e.g., pose), so the influences of different attributes on vein recognition have been poorly investigated. Therefore, this paper examines the suitability of synthetic vein images generated to compensate for the urgent lack of publicly available large-scale datasets. Firstly, we present an overview of recent research progress of palm vein recognition, from the basic background knowledge to vein anatomical structure, data acquisition, public database, and quality assessment procedures. Then, we focus on the state-of-the-art methods that have allowed the generation of vascular structures for biometric purposes and the modeling of biological networks with their respective application domains. In addition, we review the existing research on the generation of style transfer and biological nature-based synthetic palm vein images algorithms. Afterward, we formalize a general flowchart for the creation of a synthetic database comparing real palm vein images and generated synthetic samples to obtain some understanding into the development of the realistic vein imaging system. Ultimately, we conclude by discussing the challenges, insights, and future perspectives in generating synthetic palm vein images for further works.
\end{abstract}


\section{Introduction}
Biometric systems are very important in the modern era due to the increasing demand for security, using biology and technology to identify people around the world~\cite{jain2004}. Therefore, physiological and behavioral traits with properties such as universality, distinctiveness, permanence, and acceptability are used in the identification process~\cite{jain2004,wu2019}. The most commonly used characteristics are the fingerprint, iris, and face; however, they have limitations due to their exposure to the environment, making them susceptible to forgery. In contrast, more recent traits such as palm and finger veins are difficult to forge because they are not visible to the human eye~\cite{SHAHEED202284,hillerstrom}.

Palm veins become visible due to the hemoglobin in the blood plasma. For this purpose, a light array is used whose wavelengths are in the near-infrared spectrum. Thus, part of the light is absorbed by the hemoglobin, reducing the reflection range and producing gray patterns in the images, which corresponds to the vascular structures~\cite{xueyan2008,cai2010,rajalakshmi2011}. There are two methods for the acquisition of palm vein images: reflection and transmission. In the reflection method, the illumination component and the capture sensor are on the same side; these devices are low-cost and widely used in research~\cite{nikisins2010,shah2015,comert2016,kilgore2017}. Whereas in the transmission method, the illumination array and the capturing sensor are on opposite sides. Hence, the lighting intensity must be of a high magnitude to pass through the tissues and muscles of the hand, which makes the device costly~\cite{wu2019}.

Most publicly available databases use the reflection method for image acquisition. However, in the database acquisition process, there are limitations in terms of time, security, and cost, which have raised challenges to explore this technology for large-scale applications~\cite{handbook2020}. Generally, research in palm vein biometrics has been focused on verification through data augmentation per subject rather than identification since publicly available databases have a reduced number of individuals and samples, as shown in Table~\ref{t:datasets}. Thus, it is not feasible to evaluate the scalability of developed systems on massive datasets. In other biometrics such as fingerprints, handprints, or iris, the construction of large-scale synthetic databases has permitted the development of the corresponding biometric technologies. Hence, synthetic database generation allows to rapidly progress in the research field while preserving the security of users and also proving to be inexpensive compared to the traditional collection of biometric samples~\cite{crisan2008,fvc2004,palmprint,zuo2007generation}.

Regarding the image generation of vascular structures for biometric recognition, according to our literature review, there are only two approaches focused on palm vein imaging to produce synthetic ROI samples of palm vein images~\cite{icprs2021style,icprs2021nature}. Both works mark a milestone by preliminary validating and introducing two different synthetic palm vein databases, which are the largest of the state-of-the-art. However, the proposed methods are based on two generalist techniques that do not explicitly consider the topology of the palm vascular system. Moreover, other studies have provided methodologies for the generation of synthetic vein images such as dorsal hand veins~\cite{crisan2008}, finger-vein images~\cite{hillerstrom,zhang2019,yang2020gan}, and sclera vascular network~\cite{das2017}. However, it could be argued that the simulation of palm vein patterns is a difficult task due to the complex vein structures of the hand~\cite{structvein,kiss1974atlas}. In fact, the different movements of the hand require many structures working synchronously, organizing multiple vessel networks to ensure a constant blood flow.

In real images, vascular network analysis is performed using various segmentation techniques~\cite{kirbas2004survey,YIN2022199}. Particularly in~\cite{kirbas2004survey}, the methods are divided into six categories: pattern recognition techniques, model-based approaches, tracking-based approaches, artificial intelligence-based approaches, neural network-based approaches, and tube-like object detection approaches. The difference of each method lies in the ability to extract the venous pattern from the images, sharing the disadvantage that they are not robust to low-quality images, which prevents a detailed characterization of the vascular structure of the palm. Despite the above, methodologies associated with the study of different biological networks found in nature can be used to simulate the palmar venous structure. These networks have shown certain similarities with blood networks that allow their comparison~\cite{aghamirmohammadali2018, scianna, francis2009scaling, runions2005, runions2007}. Particularly, any biological network is formed by a hierarchical network composed of primary veins with branches into lower-order veins. These branches may have a free termination, producing an open venation pattern, or they may be connected by anastomosis, forming loops in a closed shape~\cite{runions2005}. Biological network modeling involves both phenomenological and qualitative parameters, such as allometric scaling laws and geometry, respectively. Besides, growth rules can be expressed by differential equations (ordinary or partial) or by computational modeling.

Extensive experimental work has been published in the literature to formulate models for the formation and growth of vein networks~\cite{scianna,prusinkiewicz2012computational,ajam2017review}. In~\cite{scianna}, the authors reviewed the mathematical models for vascular networks formation, which are based on two main mechanisms: vasculogenesis and angiogenesis. Their work emphasizes the ability to reproduce different biological systems and predict measurable quantities describing the overall processes. Additionally, a survey of mechanistic-based plant modeling techniques that approach plant morphogenesis from geometric and molecular perspectives is presented in~\cite{prusinkiewicz2012computational}. On the other hand, the authors of~\cite{ajam2017review} study modeling techniques for reconstructing blood vessels from brain images; the related models are classified into three categories: image-based modeling, mathematical methods, and hybrid models. According to their analysis, image-based modeling consists of pre-processing and feature extraction techniques, mathematical methods produce characteristics of the vascular network, and hybrid models are a fusion of the two methods. 

Contrary to previous works, the present work allows establishing state-of-the-art approaches for the generation of vascular networks to deepen knowledge towards the synthesis of palm vein images. Our objective is to determine their applicability for the simulation of different structures associated with the palm vein patterns. Additionally, we include models of networks that have been used for other purposes, such as aggregation methods that arise in non-organic environments and adaptive dynamics methods that govern the behavior of some multi-cellular organisms. Moreover, it should be noted that the works reported in the literature have focused on the generation of network structures, which only represent a single characteristic associated with the distribution of real palm vein images. However, for the generation of synthetic palm vein images with biometric purposes, it is also required to consider qualitative parameters associated with the structure, texture, illumination, and contrast. These factors are also reviewed in this paper by establishing the parameters that influence the visualization of vascular patterns. Besides, we introduce a general flowchart for creating large-scale synthetic databases of palm vein images to obtain some understanding into the development of the realistic vein imaging system.



Accordingly, the contributions of our study are as follows:
\begin{itemize}
    \item An analysis of the factors that affect the visualization of vascular structures on images, aiming to outline the parameters that must be taken into account in the development of mathematical or computational models for the generation of synthetic palm vein images, including the vein anatomical structure, the acquisition of palm vein images, quality assessment procedures, and public palm vein databases.

    \item An extensive study on the existing approaches to model vascular networks determining their applicability to produce different vein patterns that could be used to generate palm vein images.
    
    \item A general scheme for the creation of large-scale databases of synthetic palm vein images comparing real palm vein images and generated synthetic samples to obtain some understanding into the development of the realistic vein imaging system. Using this scheme, we increased the size of the proposed databases in~\cite{icprs2021style} and~\cite{icprs2021nature}, which are the largest datasets of the state-of-the-art facilitating the evaluation of large-scale biometric methods based on palm vein recognition. Moreover, the proposed scheme for image generation is incremental that allows increasing the number of individuals in the databases continuously.

    \item An overview on the main challenges, insights, and potential future research directions for the development of synthetic palm vein databases for further studies.
    
\end{itemize}

The rest of the paper is structured as follows. Firstly, Section~\ref{s:background} provides an overview of the processes involved in the visualization of palm veins, emphasizing on the vascular structure of the palm, the image acquisition process, and methods for evaluating image quality. A discussion of public palm vein databases for person recognition is given in Section~\ref{s:datasets}. Then, Section~\ref{s:review} reviews state-of-the-art approaches that provides a conceptual basis for generating synthetic images of palm veins. Later, Section~\ref{sec:results} introduces a general scheme for the creation of a synthetic database and evaluates two generalist methods for palm vein imaging. Finally, sections~\ref{s:discussion} and~\ref{s:conclusions} conclude our work by analyzing the main challenges and perspectives for further investigations on the generation of synthetic palm vein images.

\section{Background Knowledge on Palm Vein Images for Biometric Recognition}\label{s:background}

A palm vein-based recognition system can be summarized into four main processes~\cite{wu2019}: (1) image acquisition; (2) pre-processing for image segmentation and enhancement; (3) feature extraction; and (4) recognition process. Compared to other traditional biometric techniques such as fingerprint, iris, or face, the palm vein pattern as a biometric trait is relatively new; thus, there are still several challenges to study. Particularly, studying the vascular networks of the palm and the image acquisition process is essential in order to create synthetic palm vein images, which is the main aim of the present research.

\subsection{Vascular Network of the Hand}\label{ss:hand}

\begin{figure}[!htb]
  \centering
  \includegraphics[width=0.55\textwidth]{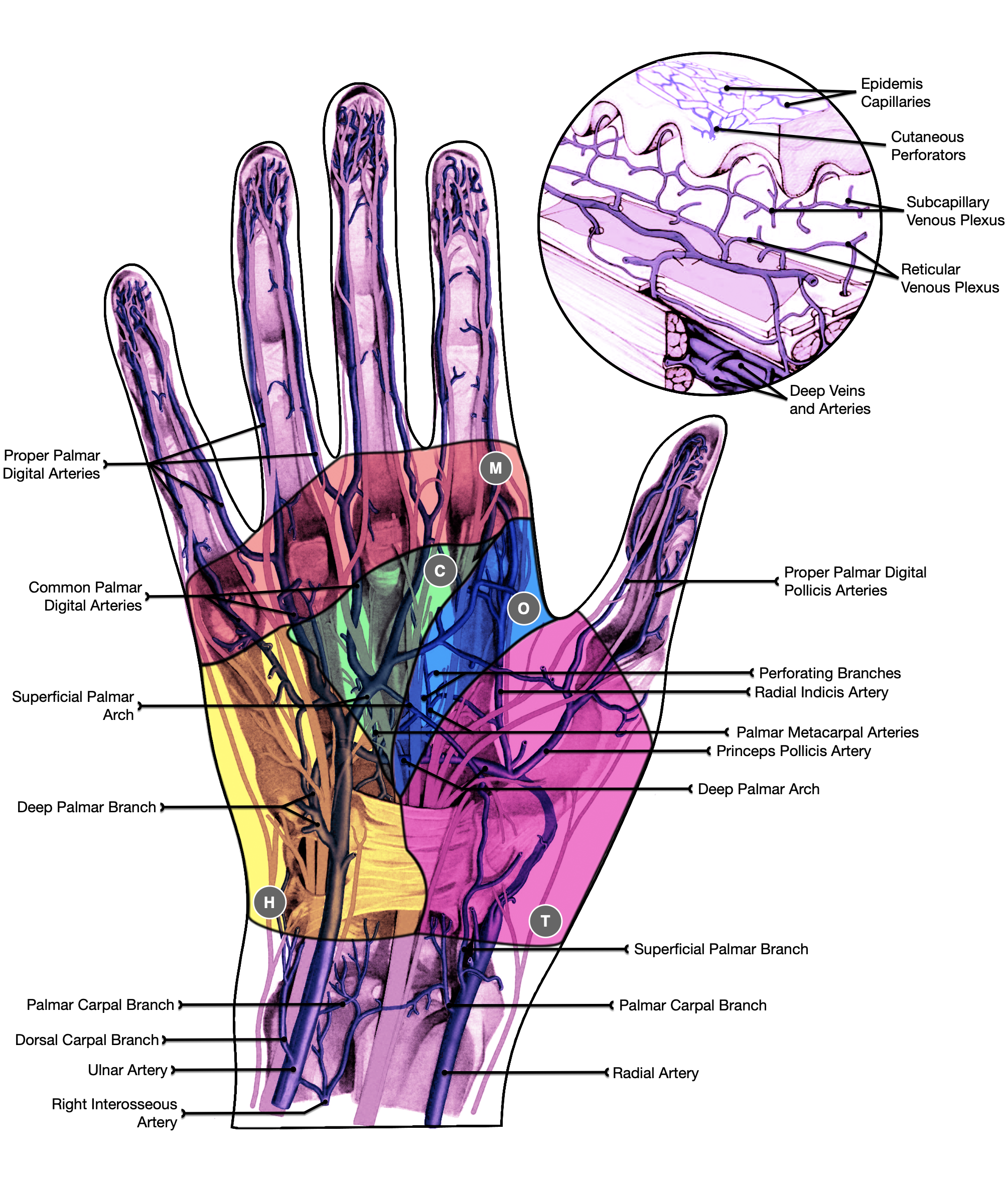}
  \caption{Representation of the vascular anatomy of the palm and functional aesthetic units: Thenar (T), Opposition (O), Hypothenar (H), Central triangular (C), and Metacarpal (M). The circular zoom shows the organization of the vascular structures inside the skin.}\label{f:palm-veins}
\end{figure}

The vascular networks of the hand are divided into veins and arteries. The veins drain deoxidized hemoglobin (Hb), and the arteries transport oxidized hemoglobin (HbO$_2$). The vascular anatomy of the palm and its functional aesthetic units are represented in Figure~\ref{f:palm-veins}, where a distinction is made between deep and superficial branches with various connections between these two sets~\cite{bergan2014}. The superficial veins are located immediately below the epidermis and between the two superficial fasciae, as can be detailed in the circular zoom of Figure~\ref{f:palm-veins}.

The vascular structure of the palm shows complex patterns that ensure unobstructed blood flow when any movement of the hand occurs~\cite{nystrom1990}. These patterns are precisely those that allow the vascular network to be a biometric trait for identification, of which superficial networks are mainly used due to the optical limitations of acquisition systems.~\cite{crisan2008}. There is a classic description of the superficial arterial topology of the palm, as shown in Figure~\ref{f:palm-veins}. A superficial palmar arch around the metacarpal bones arises from the anastomosis of the ulnar artery with the radial artery and branches into the common palmar digital arteries. In the top of the palm, where these arteries reach the proximal phalanges, they branch into palmar digital arteries to supply blood to the fingers~\cite{hansen2017}. 

\begin{figure}[!htb]
  \centering
  \includegraphics[width=0.6\textwidth]{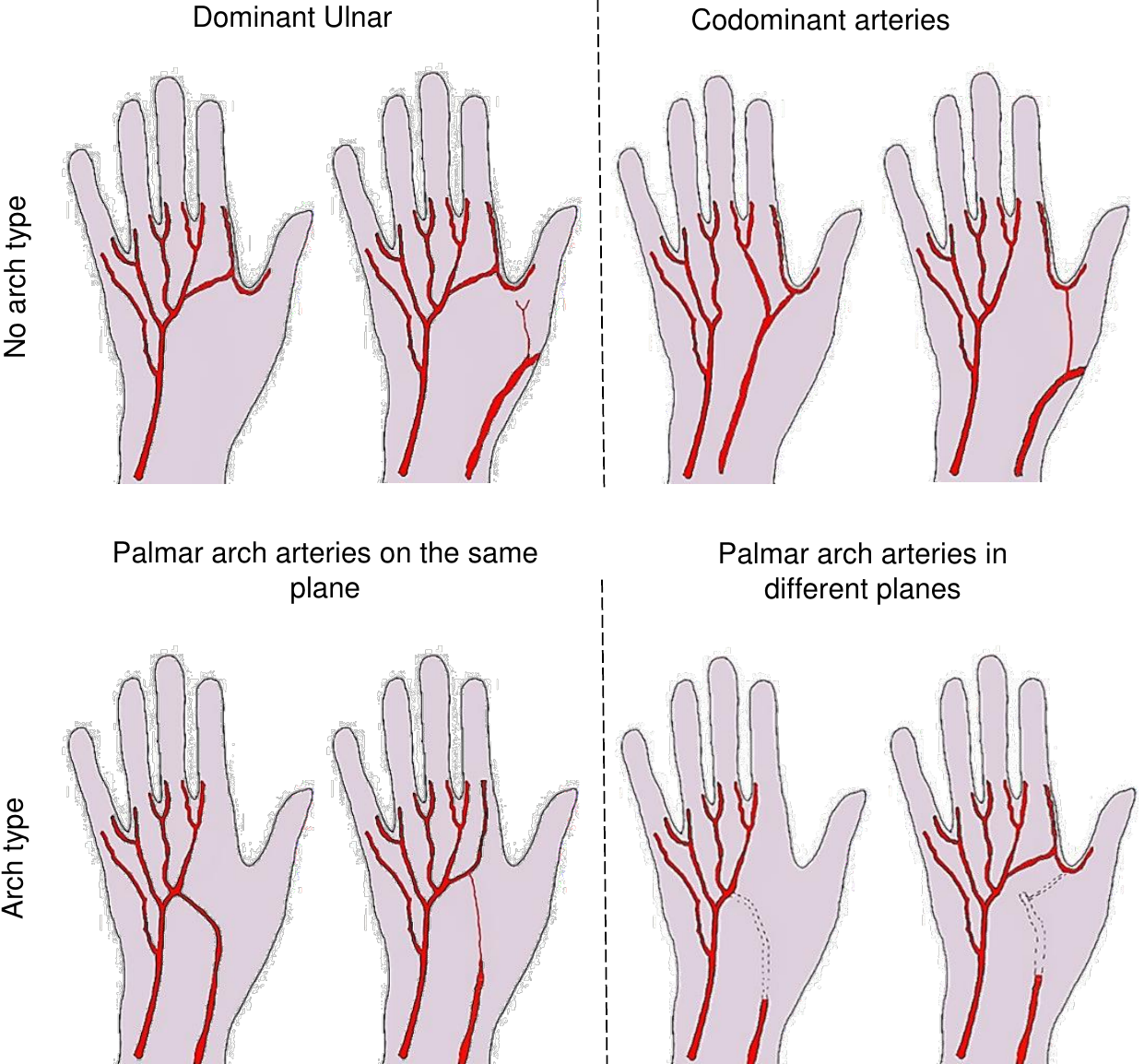}
  \caption{Structure variants of the palmar arch, based on the classification proposed in~\cite{ottone2010analysis}.}\label{f:variants}
\end{figure}

In addition to the classic structure mentioned above, there are some variants with the same relevance. According to different studies~\cite{coleman1962arterial,jelicic1988arcus,ottone2010analysis}, the typical case corresponds to only 55\% or less of the analyzed patients. Specifically, in~\cite{ottone2010analysis} they classified 86 hands according to the absence of the palmar arch or according to the anastomosis of the arteries, as shown in Figure~\ref{f:variants}. In the case of the absence of the arch, the structure may form two configurations: irrigation produced solely by the ulnar artery (Dominant Ulnar) or by two arteries independently (Codominant). The first structure was found in 29\% of the samples, of which 6\% presented a small radiopalmar artery that ends at the level of the thenar muscles (Ulnar/Radiopalmar (Thenar) Pattern).  The second case was reported in 13\%, including 8\% showed a structure formed by the ulnar and the radial artery, and 5\% comprised of the ulnar and medial arteries.

Regarding the anastomosis of the arteries that form the palmar arch, it can occur in two forms~\cite{ottone2010analysis}: the arteries forming the arch are in the same plane, or the radial artery passes through the lower part of the muscle to anatomize with the ulnar artery. In the first case, the structure was present in 43\% of the samples; 22\% corresponds to the classic case and 21\% to a variant, where the radial artery participates to a lesser extent in the arch formation (Ulnar pattern). The second case is less usual, where the network pattern only appears in 15\%; of which 7\% with a structure similar to the classic case, and the remaining 8\% is like the Ulnar pattern, the difference lies in the fact that the radial artery passes through the lower part of the muscle.

On the other hand, the topology of palm veins is a less studied topic, paying more attention to their functionality; in particular, how veins can avoid compression during grasping is studied~\cite{botte2003}. Although the structure of palm veins is unknown, research in plastic surgery has provided information about the blood supply in the skin, allowing classification of the complex vascular pathways by direct cutaneous perforator vessels and indirect cutaneous vessels~\cite{hong2018perforator,thaller2014grabb}. The former arises from the underlying veins and pass between muscles and other structures and then perforate the outer layer of the fascia where their primary destination is the skin, as shown in the zoom of Figure~\ref{f:palm-veins}. In contrast, indirect cutaneous vessels penetrate the deep tissues vertically or obliquely before perforating the outer layer of the fascia, which may emerge as small terminal branches to provide secondary supply to the skin.

Furthermore, for the location of the different perforator veins of the palm it is determined based on their five functional regions~\cite{rehim2015enhancing}: Thenar (T), Opposition (O), Hypothenar (H), Central triangular (C), and Metacarpal (M), as highlighted with different colors in Figure~\ref{f:palm-veins}. Thus, it is assumed that the direct perforants are located in the articulation of each of the regions, and the indirect perforators are within the functional regions. Therefore, each perforator forms an angiosomal module defined by a perimeter of anastomotic vessels that connect it to its neighbor in all directions.~\cite{thaller2014grabb}. The size and length of the cutaneous perforators may vary, but they all interconnect to form a network of vessels. These are well developed in the dermis, the sub-dermis, the subcutaneous's lower surface, and the deep fascia's outer surface.

\subsection{Acquisition of Palm Vein Images}\label{ss:acquisition}

\begin{figure}[!ht]
  \centering
  \includegraphics[width=0.85\textwidth]{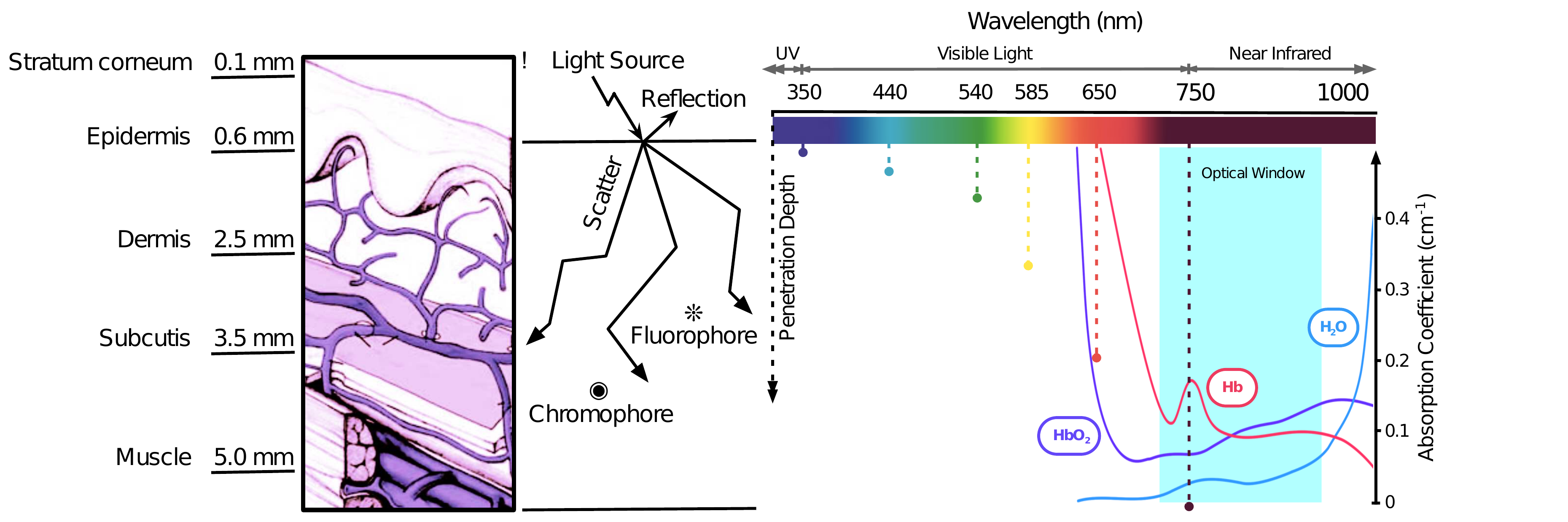}
  \caption{Penetration depth and path of the light into the skin, for different wavelengths and the corresponding absorption coefficients of oxidized hemoglobin (HbO$_2$), deoxidized hemoglobin (Hb), and water (H$_2$O).}\label{f:absorption}
\end{figure}

The acquisition of palm vein images is carried out using near-infrared (NIR) devices that interact with oxidized hemoglobin (HbO$_2$) and deoxidized hemoglobin (Hb) in the vascular network~\cite{wu2019}. Figure~\ref{f:absorption} depicts the penetration depth and path of the light into the skin and the corresponding absorption coefficients of HbO$_2$, Hb, and water. The optical window from 720nm to 950nm used for image acquisition is highlighted. When the light wavelength is between 720nm and 760nm, the radiation is strongly absorbed by Hb, producing a shadow corresponding to the vein pattern. At 790nm, there is an intersection point where Hb and HbO$_2$ present the same absorption, which allows the visualization of veins and arteries. Meanwhile, for higher spectral ranges of the optical window HbO$_2$ presents a slight increase compared to Hb.

\begin{figure}[ht!]
    \centering
    \includegraphics[scale=0.7]{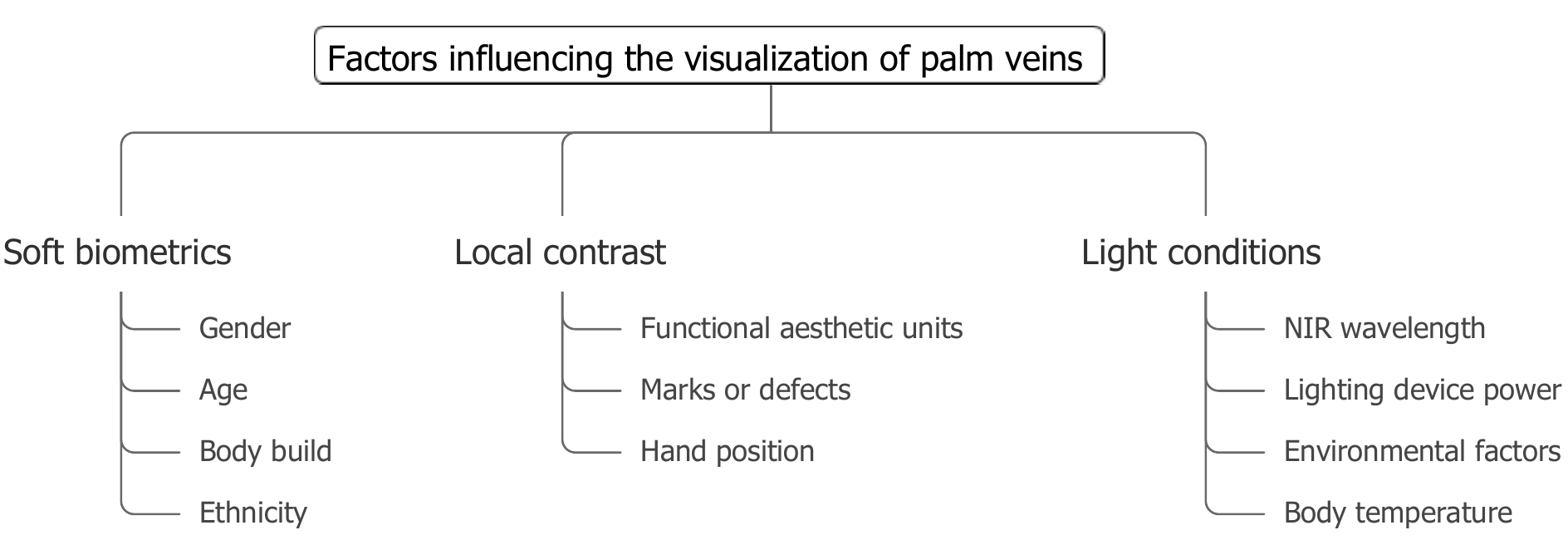}
    \caption{Classification of parameters that produce variability in the visualization of palm vein patterns in captured images.}\label{t:conditions}
\end{figure}

The different absorption rates of NIR radiation make it possible to identify the location of veins and arteries while minimizing the influence of the surrounding tissue. It is worth mentioning that the amount of optical penetration depth ranges from 0.5 to 5mm due to reflection, scattering, or fluorescence phenomena occurring in the different tissues of the palm, as shown in Figure~\ref{f:absorption}. Consequently, vein acquisition devices can only capture superficial veins~\cite{crisan2007low}. The resulting visualization of palm vein patterns in captured images depends on many variables. Figure~\ref{t:conditions} summarizes a classification of these parameters based on three groups: soft biometrics, local contrast, and illumination conditions. 


The first group of variables is related to soft biometrics, which can modify the brightness and contrast of palm vein images, such as gender, age, body build, and ethnicity. Gender can be differentiated by palm color and texture~\cite{Xie2018palmprint, Jain2004soft}. The male palm generally shows a darker partial shade and more textural differences resulting from the greater grip. Furthermore, in~\cite{Damak.etal2019palm} it is studied the features associated with vein topology, such as the gradient of the vein pattern, providing information about gender. On the other hand, the individual's age and body build affect the elicitation of the venous pattern of the palm, as the light intensity varies due to different hemoglobin and water levels in the veins and skin. The last parameter considered in this category is ethnicity since images may depend on skin tone. Particularly, in~\cite{edelman2012identification} it is studied the influence of near-infrared radiation on backgrounds of different colors. In some cases, background absorption dominates the spectra and reduces the spectral characteristics of hemoglobin. {Associated with that, it is worth mentioning that only the VERA database~\cite{VERA} collects extra information about the sex and age of the individuals. Due to the characteristics of the participants, as shown in Figure~\ref{f:VERA}, it can be noted that the representativeness of some groups is limited, which reduces the performance of gender and age classification based on palm vein images~\cite{Damak.etal2019palm,zabala2021evaluation} and limits the study of these factors on the variability of the image.}

In addition, some variables influence the local contrast, such as the functional aesthetic units of the palm, palmprint marks or defects, and the position of the hand in the acquisition device. These parameters generate regions with different levels of contrast and light intensity. The brightness and contrast in the fingers are the closest to the study of light variation in different palm regions. In~\cite{walus2017impact}, it is shown that identification results are improved using the proximal part of the finger rather than the distal part. Additionally, authors of~\cite{hillerstrom} simulate the illumination effects that occur in some regions of the finger. Particularly, they focused on the interphalangeal joints since a higher light penetration through the finger occurs because the fluid has a lower density than the bones. In the case of the palm, we can distinguish five functional regions, as shown in Figure~\ref{f:palm-veins}, which produce variations in relief and brightness in the images. This effect can be observed in the images of public datasets presented in Figure~\ref{f:datasets}, where a lower density of veins in the central regions (Opposition (O) and Central triangular (C)) can be noticed. As for the marks and defects in the palmprint being visible in the image inevitably lead to poor visibility of the vein patterns~\cite{wang2007infrared}. However, scars are skin markings that have a geometric distribution contributing to the identification of individuals~\cite{zhang2014study}. The last parameter considered concerning local contrast is the hand position, which causes geometric and perspective variations due to inconsistent positions and postures of the hands with respect to the camera~\cite{wang2016improving}. It is worth noting that~\cite{kauba2019combined} shows that tilting, bending, planar and nonplanar rotations cause severe performance degradations of publicly available datasets.

On the other hand, the illumination system is one of the most critical aspects of the image acquisition process, involving NIR wavelengths and the power of the illumination device. The NIR wavelength influences the visualization of vein patterns because of the irregularity in the absorption of hemoglobin and other tissues~\cite{walus2017impact}. Some wavelengths in the near-infrared spectrum range are more capable of acquiring biometric traits than others, as shown in multispectral databases such as CASIA~\cite{CASIA} and PolyU~\cite{PolyU}. Furthermore, the power of the illumination device must ensure uniformity and avoid over-saturation of the captured image. To regulate the illumination power, multiple layers of diffusers and different LED configurations have been used to modify the intensity distribution, mainly when the camera must be located on the same side as the light source~\cite{crisan2010radiation}. While high power light sources have the undesirable effect of decreasing contrast due to a large amount of radiation emitted, in contrast, low-intensity light sources are not able to penetrate the skin tissue, not allowing visualization of the vein pattern~\cite{Tongji}. Besides, two external parameters influence the lighting conditions, namely environmental factors and body temperature. Environmental factors such as temperature and surrounding illumination can lead to unstable gray value distribution patterns in images~\cite{heenaye2009study}. These factors acquire greater relevance when far-infrared spectroscopy techniques are used.~\cite{fan2018multiple}. Finally, body temperature is an important component of vasodilation that increases heat loss and blood flow at the palm surface, resulting in darker images~\cite{costa1998role}.

In summary, palm vein images are affected mainly because the acquisition devices are faced with uncontrolled parameters such as uneven illumination and hand position. Additionally, they are affected by device-independent parameters related to soft biometrics and lack of robustness due to outdoor illumination~\cite{IITIpalmvein, zhang2019}. To deal with the above, different acquisition devices have been proposed. Table~\ref{t:sensors} summarizes the proposed acquisition devices for palm vein imaging that change depending on the sensor type, illumination method or mounting, and hardware components, resulting in a large variety of images that cannot be compared because of the non-standardization of the method. There are some techniques to evaluate the quality of palm vein images; however, these are also not standardized~\cite{wu2019}. Therefore, considering that image quality is a key predictor in the biometric recognition process~\cite{chen2015research}, it would also allow a comparison between real and synthetic databases. The following section describes the proposed procedures for evaluating the quality of vein images.

\begin{table}[ht!]
\centering
\caption{Summary of acquisition devices for palm vein imaging.}\label{t:sensors}
\scalebox{0.77}{%
\footnotesize
\begin{tabularx}{21cm}{>{\arraybackslash}m{2.3cm} >{\arraybackslash}m{3.7cm} >{\arraybackslash}m{4cm} >{\centering\arraybackslash}X >{\arraybackslash}m{6.5cm}} 
\toprule
\textbf{Sensor Type}  & \textbf{Method and NIRline Spectrum} & \textbf{Hardware andline Components}  & \textbf{Contactless Device}  & \textbf{Pros and Cons} \\
\midrule
 Multispectral CCD line Camera~\cite{CASIA}   & Multispectral Reflection (460nm, 630nm, 700nm, 850nm, 940nm, VIS)  & Infrared LEDsline (violet to IR), optical diffusers    & Yes & The recognition is performed both with the palm print and the internal structures. However, the vein pattern is barely visible. 
 \\ \hline
 ImagingSource Sony ICX618~\cite{VERA} & Reflection (940nm) & HC-SR04 ultrasound sensor and a led signal (palm position)  & Yes & The system is sensitive to the position of the hands, sometimes generating blurring images. 
 \\ \hline
 USBline camera~\cite{PUT} & Reflection (880nm) &  Infrared LEDs (880 nm), LED diffusers   &  No &  It is a contact acquisition device, which allows controlling the illumination conditions to obtain a better vein pattern. However, it reduces the level of hygiene.\\ 
 \hline
  Multispectral CCDline Camera~\cite{PolyU}   & Multispectral Reflection (470nm, 525nm, 660nm, 880nm)  & Multispectral light, standard closed-circuit television camera, light controller, monochromatic CCD & No & The multispectral range allows highlighting different texture information of the palm. However, it is not clear how to best merge multispectral information. \\
 \hline
  CCDline camera~\cite{Tongji} & Reflection (940nm)   & Ring near infrared LED light source, a power regulator, LCD with a touch screen & Yes  & It is possible to adjust the illumination power and determine the palm position on the LCD screen to obtain a good image. However, the device has the disadvantage that it is not automated. \\
 \hline
 USB camera mvBluefox IGC~\cite{IITIpalmvein} & Reflection (NIR array 850nm and 940nm)  & Infrared LEDs array (850nm and 940nm) & Yes &  Infrared
LEDs were placed alternately. Three arrays of LEDs were used: two of them to illuminate the hand, and the third to illuminate the fingers. The finger-vein information can improve the performance of the palm vein recognition. \\ 
 \hline
 Adaptedline webcam~\cite{Tecnocampus} & Multispectral Reflection (VIS, 700nm, Thermal)  & Thermal camera, webcam, filters for IR, infrared LEDs & No & The device allows the capture of 3 images: in the visible spectrum (VIS), IR spectrum, and thermal.  The results show that VIS images provide the best identification rates. The above indicates that the IR setup is not of good quality. \\
\hline
CMOSline camera~\cite{FYO} & Reflection (IR)  & Camera with adjustable support (360$^o$ rotation in the plane of the palm) and wooden hand guides & No & CMOS sensors employ a smaller digital circuit that uses less power and are less expensive than other devices.However, the images produced have lower visual quality. \\
\hline
Adaptedline webcam~\cite{michael2010design} & Reflectionline (900nm to 920nm) & Infrared LEDs (880nm to 920 nm), optical diffusers & Yes & The user is not restricted to a particular position because the image is saved as a real-time video sequence at 30 fps. However, the storage device may require a lot of memory \\
\hline
 CCD camera~\cite{zhang2010palm} & Reflection (850nm) & Concentric LED arrays, holographic diffusers, lighting control system & No &  The diffuser scatters the light diminishing the radiation intensity. \\
\bottomrule
\end{tabularx}%
}
\end{table}

\subsection{Image Quality of Palm Vein Images}

From a general perspective, the quality assessment of infrared images has been performed mainly by human inspection using the National Imagery Interpretability Rating Scale (NIIRS) standard~\cite{irvine1997national}. The aspects involved in image interpretability are resolution, contrast, and edge sharpness. For biometric images such as fingerprints, iris, and face, there is the ISO/IEC 29794 standard~\cite{ISO4,iso1,iso5} that takes into account geometric and posture aspects. Particularly, for palm vein recognition as a contactless biometric system, the image quality greatly reduces the identification performance. However, in the case of vascular-based biometrics, there is still no standardization that provides a metric to assess the image quality~\cite{wu2019}. Consequently, each research related to the quality of palm vein imaging uses self-created metrics leading to the respective quality scores. To the best of our knowledge, Table~\ref{t:calidad} summarizes the published studies in this area, which have been limited to working with self-constructed images. Following, we describe the main characteristics of each work.

\begin{table*}[htb]
\centering
\caption{Reported studies on the image quality assessment of palm vein images. Evaluation range, lowest and highest scores are from the referenced papers; N/A means "not applied" or "not reported".}\label{t:calidad}
\scalebox{0.9}{%
\footnotesize
\begin{tabular}{m{3cm} m{4cm} m{4.5cm} m{1.7cm} m{1.3cm} m{1.3cm}}
\toprule
\textbf{Method} & \textbf{Parameters of the acquisition system} & \textbf{Evaluation parameters} & \textbf{Evaluation range} & \textbf{Lowest score} & \textbf{Highest score} \\ \midrule
High quality palm vein image acquisition~\cite{chen2015research} & Intensity of NIR light source and camera parameters & 2D image entropy and local 2D entropy & N/A & N/A & N/A \\ \hline
Image quality and recognition with different distance~\etal~\cite{jiaqiang2013analysis} & Distance to the camera lens & Image clarity, image structural difference, and match results & 0 -- 1 & N/A & N/A \\ \hline
Image quality assessment based on tilt angle~\cite{wang2017quality} & Tilt angle of the palm ($0^o$ -- $50^o$, with step of $10^o$) & Clarity and brightness uniformity & 0 -- 1 & 0.446 & 0.543\\ \hline
Multi-spectral adaptive capturing system based on image quality~\cite{dong2017research} & Intensity of the multi-spectral light source & Mean and variance of gray levels & 0 -- 1 & 0.3 & 0.8\\ \hline
Palm vein image quality assessment by using natural scene statistics features~\cite{wang2017statistics} & Tilt angle of the palm ($0^o$ -- $50^o$, with step of $10^o$) & Illumination uniformity, natural image quality evaluator (NIQE), and statistics of Harr-Like features & N/A & 4.190 & 1.622 \\ 
\hline
 GLCM metrics~\cite{michael2010design} & N/A & Threshold of acceptance as a function of contrast, variance, and correlation & N/A & N/A & N/A \\
\bottomrule
\end{tabular}%
}
\end{table*}

In~\cite{chen2015research}, a device for acquiring high-quality images of palm veins is proposed to avoid the low quality of captured images. The quality of the images is evaluated using the 2D image entropy and the local 2D entropy. The calculation is performed by the accelerated implementation method using Field Programmable Gate Array (FPGA) platform. The acquisition device controls the intensity of the NIR light source and camera parameters based on the image quality. Therefore, taking into account the quality of the resulting image, the device recaptures the image until a high-quality image is obtained.

On the other hand, the authors of~\cite{jiaqiang2013analysis} evaluated the image quality as a function of the distance between the palm to the camera lens. The paper studies the relation between distance and recognition performance based on the image quality evaluation. The proposed evaluation method consists of two parts: (1) analyzing the detail change from the image clarity and (2) examining the structural difference of the image. The first part is performed through the Tenengrad energy gradient function, which uses the Sobel operator to extract the horizontal and vertical gradient values. The second part is performed by using the SSIM structural difference method~\cite{wang2004image}, which is calculated as a function of light, structure, and contrast. Although this work shows that the image quality depends on the distance from the palm to the camera, they do not specify the ranges in which the image quality oscillates. Besides, their results determine an adequate distance range for the acquisition of the vein pattern; outside these values, the false rejection rate increases.

Subsequently, Wang~\etal~\cite{wang2017quality} introduced an algorithm to evaluate the quality of palm vein images based on the tilt angle of the hand and a fusion of two criteria, namely clarity and brightness uniformity. The image clarity is estimated based on the method developed in~\cite{crete2007blur}, by performing discrimination between different levels of image blurring. On the other hand, the brightness uniformity is computed by averaging the gray difference between different parts of the palm vein image. Both criteria are combined by using a weighted quality score in the range $[0..1]$, where 1 corresponds to the best quality image. The evaluation of the proposed algorithm was performed on 60 images of different individuals by varying the tilt angle and blurring from the same person's hand. Finally, they found that the image quality decreases as the tilt angle increases; meanwhile, it is less sensitive to brightness uniformity. In their experiments, the quality score ranged from 0.446 to 0.543. However, it should be noted that more extensive experimentation should be conducted.

Moreover, Dong~\etal~\cite{dong2017research} designed a multi-spectral adaptive device to examine the image quality as a function of light source intensity. The proposed capturing system allows adjusting three light source wavelengths (760nm, 850nm, and 940nm) and then collecting the high-quality palm vein image. The image quality index is calculated by integral scores from the mass fractions of mean and variance gray levels. Their analysis of experimental results showed that variations of illumination power produce images with different qualities, ranging from 0.3 to 0.8. The overall evaluation range is $[0..1]$, where 1 represents the best quality. As in previously discussed works, the conducted experimental assessment is quite preliminary.

Lastly, to our best knowledge, the most complete work reported is~\cite{wang2017statistics}. Authors propose a methodology to evaluate palm vein image quality based on statistics of natural scenes and reject poor-quality images. In this way, the image quality is the result of three quantitative measures. The first proposed measure computes the difference between the minimum and the maximum gray mean to estimate the brightness uniformity among several image patches. Secondly, statistics of pixel intensities are calculated by using the natural image quality evaluator (NIQE) method proposed in~\cite{mittal2012making}. For the last metric, statistics of four kinds of Harr-Like features are computed by using natural scene statistics and the Mahalanobis distance. The overall image quality score is equal to the brightness uniformity score times a linear combination of the other two measures. The evaluation of the algorithm was performed on 60 self-captured images from 21 individuals. Experimental results show the effect of uneven illumination due to different tilt angles of the palm and the quality scores for different distortion levels. The image quality as a function of hand position ranged from 4.190 to 1.622, and as a function of distortion ranged from 8.605 to 2.261, where the lower value corresponds to the best quality image. Despite this, the quality score is not bound in a closed range.

Furthermore, in order to broaden the scope of quality assessment methods of palm vein images, related works on finger-vein imaging have also been considered as the acquisition devices share many similarities~\cite{hillerstrom}. In~\cite{ma2012finger} they proposed an extension of the Peak-Signal-to-Noise (PSN) index that incorporates properties of the human visual system and measurements of finger-vein imaging such as finger area and displacement. Besides, Yin~\etal~\cite{yin2011sdumla} utilized the ability to filter out low-quality finger vein images. On the other hand, authors of~\cite{qin2017deep} and~\cite{qin2015finger} evaluated the image quality by employing the binary segmentation of vessel structure. The former method is based on support vector regression, and the latter one uses a CNN approach. As supervised learning methods, both approaches share the disadvantage of requiring a significant amount of labeled training data.  

In our understanding, the evaluation of the quality of palm vein images has only been performed on self-captured images. Consequently, there are no common elements to compare quality between different public databases. Therefore, there is no evidence of intra-class and inter-class variability for the generation of synthetic palm vein image databases.

\section{Publicly Available Datasets on Palm Vein Images}\label{s:datasets}

Among all the real-world applications, palm vein recognition has been adopted for authentication purposes in many kinds of public uses~\cite{shinzaki2020palm-applications}. Fujitsu Labs provides a palm vein scanner and SDK for palm vein authentication, which reports a very low False Acceptance Rate (FAR) of 0.00008\% and a False Rejection Rate of 0.01\%~\cite{FUJITSU}. In this context, Fujitsu conducted the most extensive research study on palm vein recognition with a database comprised of 75,000 subjects and a total of 150,000 palm vein images. However, their technology is private and patented, not allowing the use of captured palm vein images to evaluate different approaches or create public datasets of palm vein patterns for research purposes. Moreover, the study was conducted for commercial purposes, and the database is not available to reproduce the experiments or compare new approaches.

\begin{figure*}[ht!]
    \centering
    \includegraphics[width=0.95\textwidth]{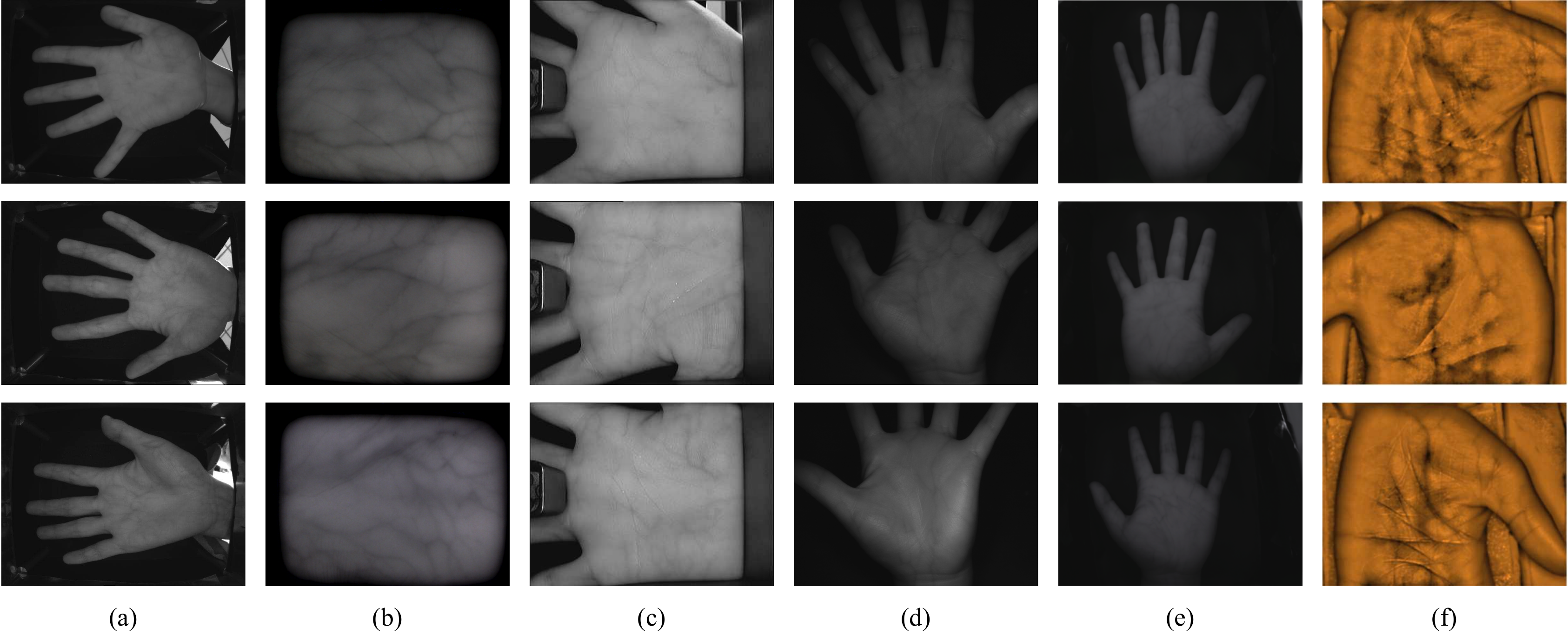}
    \caption{Exemplary of original palm vein samples from the public databases: (a) CASIA, (b) PUT, (c) PolyU, (d) Tongji, (e) IITI, and (f) FYO. Images from the VERA and Tecnocampus datasets are not shown due to restrictions of their terms of use.}\label{f:datasets}
\end{figure*}

In this context, the research community has introduced some self-created databases of palm vein images~\cite{handbook2020}, which have been collected by using self-designed and low-cost devices. As a result of the above, many non-standardized approaches use different collection protocols and different acquisition devices and equipment. Besides, these conditions produce different image parameters, variations in illumination and contrast, and also performance results. Table~\ref{t:datasets} provides an overview of the publicly available databases, comparing the number of subjects, captured hands and samples, acquisition sessions, total images, and their resolution. Additionally, Figure~\ref{f:datasets} shows image samples from each dataset.

\begin{table*}[ht!]
\centering
\caption{Details of the publicly available databases of palm vein images. The number of subjects and the number of samples are expressed as Subjects $\times$ Hands, and Samples $\times$ Acquisition Sessions, respectively.}\label{t:datasets}
\scalebox{0.95}{%
\footnotesize
\begin{tabular}{cccccccm{3.7cm}}
\toprule
\multirow{2}{*}{\textbf{Database}} & \multirow{2}{*}{\textbf{Subjects}} & \multirow{2}{*}{\textbf{Samples}} & \textbf{Total} & \textbf{Image} & \textbf{Sessions} & \multirow{2}{*}{\textbf{Labeling}} & \textbf{Dataset} \\
 &  &  & \textbf{Images} & \textbf{Size} & \textbf{Interval} &  & \textbf{Available}  \\
\midrule

{CASIA~\cite{CASIA}} & {100 $\times$ 2} & {3 $\times$ 2} & {7,200} & {768 $\times$ 576} & {30 days} & {Id} & \href{http://biometrics.idealtest.org/\#/datasetDetail/6}{National Laboratory of Pattern Recognition (NLPR)} \\
\hline
\multirow{2}{*}{VERA~\cite{VERA}} & \multirow{2}{*}{110 $\times$ 2} & \multirow{2}{*}{5 $\times$ 2} & \multirow{2}{*}{2,200} & \multirow{2}{*}{480 $\times$ 680} & \multirow{2}{*}{5 minutes} & Id, Age, &  \multirow{2}{*}{\href{https://www.idiap.ch/en/dataset/vera-palmvein}{Idiap Research Institute}} \\
 &  &  &  &  &  & Gender &  \\ 
\hline
PUT~\cite{PUT} & 50 $\times$ 2 & 4 $\times$ 3 & 1,200 & 1280 $\times$ 960 & 7 days & Id & \href{https://biometrics.cie.put.poznan.pl}{Poznan University of Technology}  \\
\hline
{PolyU~\cite{PolyU}} & {250 $\times$ 2} & {6 $\times$ 2} & {6,000} & {352 $\times$ 288} & {9 days} & {Id} & \href{https://www4.comp.polyu.edu.hk/~biometrics/}{Biometrics Research and Innovation Centre (BRIC)} \\
\hline
Tongji*~\cite{Tongji} & 300 $\times$ 2 & 10 $\times$ 2 & 12,000 & 800 $\times$ 600 & 61 days & Id & By e-mail requesting \\
\hline
{IITI*~\cite{IITIpalmvein}} & {185 $\times$ 2} & {6 $\times$ 1} & {2,220} & {2592 $\times$ 1944} & {N/A} & {Id} & By e-mail requesting  \\
\hline

{Tecnocampus~\cite{Tecnocampus}} & {100 $\times$ 2} & {2 $\times$ 5} & {6,000} & {640 $\times$ 480} & {28 days} & {Id} & \href{http://splab.cz/en/download/databaze/tecnocampus-hand-image-database}{Signal Processing Laboratory (SPLab)} \\
\hline
FYO~\cite{FYO} & 160 $\times$ 2 & 1 $\times$ 2 & 640 & 800 $\times$ 600 & 10 minutes & Id & \href{https://fyo.emu.edu.tr/en}{Eastern Mediterranean University} \\
\hline
Synthetic-sPVDB~\cite{icprs2021style} & 10,000 & 6 $\times$ 1 & 60,000 & 128 $\times$ 128 & N/A & Id & \href{https://www.litrp.cl/repository}{LITRP Lab} \\
\hline
NS-PVDB~\cite{icprs2021nature} & 2,000 & 6 $\times$ 1 & 12,000 & 128 $\times$ 128 & N/A & Id & \href{https://www.litrp.cl/repository}{LITRP Lab} \\
\bottomrule
\multicolumn{8}{l}{\small{*These datasets are publicly available by requesting to the principal investigator.}}
\end{tabular}%
}
\end{table*}

As can be noticed in Table~\ref{t:datasets}, public databases are usually composed of a limited number of people with no more than 300 individuals. Usually, the proposed approaches on palm vein recognition follow the \textit{nom L\&R} protocol from~\cite{VERA}, where the left and right hands of the same subject are considered different subjects, doubling the number of subjects in the datasets. However, there is still a lack of increasing the number of individuals in the datasets to allow the study of scalability of the proposed methods for massive identification, \ie~at least up to a city population of hundreds of thousands to millions of people. A possible solution to address this drawback could be to use images from all databases, but this would not solve the problem because of two main reasons. First, only a total of 2,510 individuals would be obtained, which would still be insufficient to evaluate methods for massive recognition of people. Second, and most noteworthy, the characteristic variations of images resulting from different acquisition systems (\eg~resolution, illumination, contrast, etc.), as shown in Figure~\ref{f:datasets}, may cause problems in training and parameter tuning of the proposed algorithms.

During the last years, the most recent approaches on palm vein recognition have focused on deep neural networks (DNN) (\ie~convolutional neural networks (CNN)), which are able to learn more robust feature representations. DNN-based methods have improved the state-of-the-art (SOTA) results on palm vein databases, achieving 100\% of accuracy for four of eight public databases (\ie~Tongji~\cite{Tongji}, VERA~\cite{VERA}, PolyU~\cite{PolyU}, and FYO~\cite{FYO}). However, DNN-based models have serious issues related to available training datasets, adjusting many parameters, scalability, among others. In this context, the small number of samples per class restricts the available training data for vein recognition tasks, which, at the same time, reduces the generalization performance because DNN is highly dependent on the number of training samples. The above issues have motivated two main solutions: the use of pre-trained models and the application of data augmentation techniques or complex training processes for a task-specific network. Besides, it is important to note that the dataset partitions (\ie~training, validation, and testing) are not standardized, even among the existing approaches themselves~\cite{jia2020performance}, which limits the comparison of the proposed approaches.

\begin{figure}[ht!]
\centering
\begin{tikzpicture}
\begin{axis}[
width=0.5\textwidth,
xbar,
xmin = 0, 
xmax = 80,
enlarge y limits=0.2,
legend pos=south east,
xlabel={number of individuals},
ylabel={age groups},
symbolic y coords={18-27,28-37,38-47,48-60,All},
ytick=data,
nodes near coords,
nodes near coords align={horizontal},
style = {font=\scriptsize},
]
\addplot[
            blue!70,
            fill=white,
            pattern=north east lines,
            pattern color=.,                
        ]  coordinates {(30,18-27) (14,28-37) (14,38-47) (12,48-60) (70,All)};
\addplot[
            red!70,
            fill=white,
            pattern=north west lines,
            pattern color=.,                
        ] coordinates {(24,18-27) (7,28-37) (6,38-47) (3,48-60) (40,All)};
\legend{Male,Female}
\end{axis}
\end{tikzpicture}%
\caption{{Overall distribution of age and sex groups from the VERA database.}}
\label{f:VERA}
\end{figure}
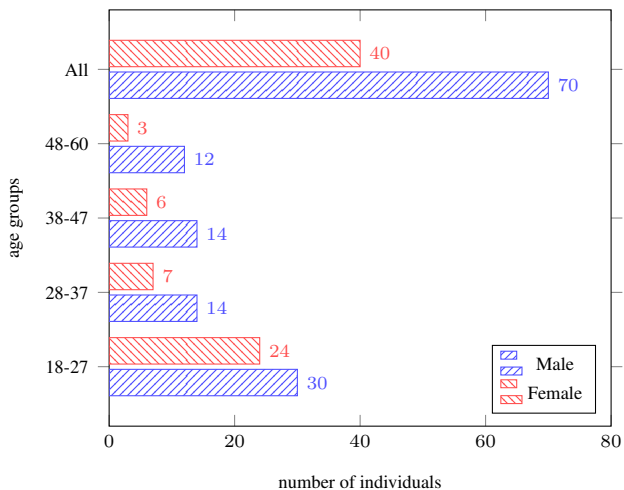

Additionally, it is relevant to notice that only the VERA Palm vein dataset~\cite{VERA} includes additional labeling metadata such as gender and age. {The database comprises 40 women and 70 men whose ages are between 18 and 60, with an average of 33 years old. Figure~\ref{f:VERA} shows the overall distribution of age and sex groups from the VERA database. It is noteworthy that there are significant differences between groups, lacking more representativeness of data to conduct deepen studies.} The extra labeling metadata is a piece of valuable information because it is demonstrated that vein patterns are affected depending on some soft biometric traits such as gender, ethnicity, physiological composition, and medical conditions~\cite{Xie2018palmprint, Jain2004soft, Damak.etal2019palm, zabala2021evaluation, finger-vein-review2018}. Thus, this auxiliary information can contribute to improving the accuracy of the recognition process. 

To address the aforementioned limitations, some biometric techniques (\eg~fingerprint, iris, or face) have adopted the generation of synthetic databases. Particularly, synthetic images are very effective for evaluating the performance of several image processing algorithms, but their use for biometric purposes is a controversial issue. However, synthetic databases have the advantage of avoiding a time-consuming collection process and also do not compromise users' security at all, which is usually a personal concern regarding privacy violations~\cite{fvc2004,palmprint,zuo2007generation}. Moreover, this kind of dataset does not entirely replace the validation with real images, whereas the generation of synthetic samples facilitates the validation of proposed approaches on large-scale databases.

Thereby, in our previous works~\cite{icprs2021style,icprs2021nature}, we proposed two synthetic databases of palm vein images based on two generalist approaches. The first dataset, named Synthetic Style-based Palm Vein Database (Synthetic-sPVDB), comprises 10,000 individuals with six samples each, totaling 60,000 images. The generation method was implemented using the StyleGAN model~\cite{karras2019style}. The other dataset, called Natural-based Synthetic Palm Vein Database (NS-PVDB), was created using a procedure based on biological transport networks~\cite{liu2017new}, being comprised of 12,000 images corresponding to 2,000 different subjects. The generated datasets are the largest in the state-of-the-art and were validated using qualitative and quantitative metrics to measure their similarities compared to the real images. 
As part of the present work, and taking into account the previous results, in Section~\ref{sec:results} we present the formalization of a general flowchart for the generation of large-scale synthetic databases of palm vein images. Besides, we introduce larger versions of the proposed databases and extend experimental results in order to validate them.

Then, the present study aims to analyze and systematize the information available in the state-of-the-art regarding the generation of synthetic images of venous structures. Most studies in the literature related to the synthesis of the human vascular network have been oriented to medical purposes either to describe the growth of tumors or to diagnose diseases. On the other hand, within the biometric context and according to the bibliographic review, there are few works focused on providing methodologies for the generation of synthetic images for vein-based recognition, such as finger-vein~\cite{hillerstrom}, dorsal-hand~\cite{crisan2008}, and sclera~\cite{das2017}. Hence, in the following sections, we review the most relevant state-of-the-art approaches, discussing their advantages and disadvantages towards the challenging task of synthetic palm vein imaging for biometric purposes.

\section{Review on the Generation of Synthetic Images of Vascular Networks Patterns}\label{s:review}

The lack of large-scale databases of vein patterns images has greatly limited the development of large-scale recognition methods that can be implemented for massive individuals identification~\cite{hernandez2019massive, hernandez2019individuals, guidet2020exhaustive}. In other biometric domains, such as fingerprints, have benefited from synthetic images that have avoided all issues concerning privacy regulations, allowing the rapid development of biometric technologies~\cite{handbook2020}. As for synthetic vein imaging, it faces issues such as low image quality and complex vascular patterns formed~\cite{botte2003}. Aiming to deepen knowledge towards the synthesis of palm vein images, we review state-of-the-art models that have allowed the generation of vascular structures for biometric purposes and the modeling of biological networks with their respective applications.

\subsection{Generation of Vein Networks Oriented to Biometric Recognition}\label{ss:RedesBiometria}

To the best of our knowledge, there are few works focused on generating synthetic vein images for biometric purposes. Table~\ref{t:synthetic-biometric} summarizes the reported approaches for different biometric purposes such as the synthesis of dorsal hand veins~\cite{crisan2008}, finger-vein images~\cite{hillerstrom,zhang2019,yang2020gan}, sclera vascular network~\cite{das2017}, and the ROI of palm vein images~\cite{icprs2021style,icprs2021nature}. 

\begin{table}[ht!] 
\centering
\caption{Overview of state-of-the-art works on the generation of vein networks oriented to biometric recognition. The database size is given in terms of number of generated images; N/A means "not applied" or "not reported".}\label{t:synthetic-biometric}
\scalebox{0.95}{
\footnotesize
\begin{tabular}{m{8cm}m{3cm}m{2.5cm}m{1.7cm}}
\toprule
\textbf{Method} & \textbf{Biometric Trait}  & \textbf{Imaging Task} & \textbf{DB Size}  \\ \midrule
Vascular structure using key points and image texture based on the coefficient of contrast variation~\cite{crisan2008} & Dorsal hand veins & Synthesis & 1,000 \\ \hline
Finger vascular structure using a growth method inspired by the generation of venation patterns of plant leaves
~\cite{hillerstrom} & Finger vein & Synthesis & 50,000 \\ \hline
Finger-vein image augmentation based on the data distribution of real images by using GAN networks~\cite{zhang2019} &  Finger vein & Augmentation & N/A \\\hline
Vascular structure based on anatomy and image translation based on CycleGAN~\cite{yang2020gan} & Finger vein & Synthesis & 53,630 \\\hline
Generation of images using a non-parametric texture algorithm~\cite{das2017} &  Sclera vessel & Synthesis &  N/A \\\hline
Application of the style-based GAN architecture (StyleGAN) for generating synthetic ROI of palm vein images~\cite{icprs2021style} &  Palm vein & Synthesis &  60,000 \\\hline
A nature-based method (inspired by the growth of the Physarum polycephalum) for the generation of synthetic ROI of palm vein images~\cite{icprs2021nature}
&  Palm vein & Synthesis &  12,000 \\
\bottomrule
\end{tabular}}
\end{table}

According to our review, the first work associated with synthetic imaging of vascular structures for biometric recognition appears in~\cite{crisan2008}. The proposed simulation method starts from randomly but realistically placed nodes\footnote{The dorsal hand vein simulator is available at \url{http://veinsim.maditech.ro/}}. Each point in the model is classified as termination, node, or a point belonging to a segment. Once the main points and the number of neighbors are obtained, an intermediate model is explored following some vital rules. After calculating the segments, a dilation algorithm is used to recreate the thickness of the veins. The maximum veins' width is calculated based on the overall dimensions of the simulated hand or entered as a constant. Finally, they recreate the real texture by calculating the coefficient of contrast variation and including some details such as virtual hair. Their findings report that 1,000 generated images are similar to real images but do not provide information on any quantitative or qualitative test results.

In the research proposed by Hillerstr\"om~\etal~\cite{hillerstrom}, they developed a methodology to generate synthetic finger-vein images\footnote{The finger vein simulator is available at \url{https://web.comp.polyu.edu.hk/csajaykr/fvgen.htm}}. The proposed model is divided into three parts.  First, venous nodes are generated using a growth method inspired by the generation of venation patterns of plant leaves~\cite{runions2005}. For this purpose, the procedure takes into account the anatomy of the finger vascular structure, making the distinction of two principal veins on both radial and ulnar sides~\cite{structvein}. The second part is the generation of the vein patterns from the ganglia, connected with a thickened pattern that characterizes the vascular network by dilatation and erosion. The vein thickness is calculated using Murray's law\cite{lee2010}, which states that the cube of the diameter of a principal vessel is equal to the sum of the cubes of the diameters of secondary vessels. Finally, the process incorporates the influence of illumination, taking into account parameters associated with the attenuation of transmitted light, optical blur, and scattering. For the paper, the authors generated a synthetic database consisting of 5,000 different subjects with 10 synthetic finger-vein images per each, for a total of 50,000 finger-vein images, which is the largest dataset of finger-vein images. To verify the similarity between synthetic images and real images, they use the Bessel K-forms~\cite{grenander2001probability} resulting in quite realistic results. Despite this, their database has not yet been used by other studies.

With the rise of Generative Adversarial Networks (GANs), due to their excellent performance for synthetic imaging~\cite{jabbar2021gansurvey,SHAMSOLMOALI2021126}, some authors have adopted GANs architectures to generate finger-vein images~\cite{zhang2019,yang2020gan}. Zhang~\etal~\cite{zhang2019} introduced a lightweight GAN architecture to implement finger-vein image augmentation, called FCGAN, which uses a preliminary batch normalization and a tightly-constrained loss function. Their proposal aims to improve Deep Learning methods for finger-vein recognition by reducing training overfitting. Results from experiments on sample augmentation show that CNN converges faster when trained using FCGAN-augmented samples. Hence, it means that FCGAN is able to generate diverse and high-quality finger-vein images, which is useful for improving the training procedure of CNN-based models. Besides, from a visual-quality point of view, the grid effect and the diversity of synthetic samples were analyzed. By using the FCGAN architecture, the strip-shaped creases do not appear in the finger-vein images, generating smoother images. Moreover, FCGAN generates synthetic images with more uneven tones, increasing the diversity of samples. Nevertheless, it should be noted that the proposed approach has only been used for image augmentation rather than creating new samples of different individuals for synthetic database creation.

On the other hand, in~\cite{yang2020gan}, they constructed a synthetic dataset of 53,630 finger-vein images corresponding to 5,363 different subjects. For the generation of the images, they first develop the synthetic vein pattern using the algorithm developed in~\cite{runions2005} and subsequently perform the translation to images using the CycleGAN architecture~\cite{zhu2017unpaired}. The model learns a correct mapping $X \to Y$ through a generator, where $X$ contains the data distribution of vascular structure and $Y$ represents the domain containing the image of the structure. Besides, CycleGAN implements a second generator to learn the mutual transformation $Y \to X$. In order to minimize the difference between the synthetic vascular structure and the real image structure, they use the approach proposed by Johnson~\etal~\cite{johnson2016perceptual} in the generator. Instead of using a fully-connected neural network as a discriminator, authors used $8 \times 8$ PatchGAN~\cite{isola2017image}, which is able to process images of arbitrary size and has a smaller number of parameters. For evaluation, they trained the model with ROI images of the MMCBNU\_6000 finger-vein database~\cite{MMCBNU} and validated the synthetic dataset by using five different metrics. Compared against real finger-vein images, experimental results show the generated dataset is very close to the real data distribution, and also venous patterns are better highlighted. 

On other biometric domain, Das~\etal~\cite{das2017} focused on the sclera vessel synthesis to improve the robustness of ocular biometric systems. To produce the synthetic images of the sclera, they develop an algorithm based on non-parametric texture regeneration proposed by~\cite{efros1999}. The method consists of modeling the texture as a Markov random field. Thus, given the brightness values of the spatial neighborhood, it is assumed that the brightness probability distribution for a pixel is independent of the rest of the image. Initially, the vein pattern texture is initialized from the source texture with a $3 \times 3$ pixel seed. For unfilled pixels, the proposed technique finds a set of patches that most closely resembles the filled neighbors of the unfilled pixel in the source image. One of these patches is randomly chosen, and a color value is assigned to the unfilled pixel from the center of the selected patch. The above process is repeated until the texture is obtained. The experiments were carried out by using the UBIRIS v1 database~\cite{proencca2005ubiris} manually cropped as primitive samples, and they generated a synthetic image for each real one. The proposed experimental setup evidences the usefulness of the generated sclera patterns for biometric purposes. However, it is noteworthy that their implementation is very time-consuming.

Regarding the generation of synthetic palm vein images, it is relevant to highlight our previous conference works presented in~\cite{icprs2021style,icprs2021nature} and extended in Section~\ref{sec:results}. These studies preliminary evaluate two generalist techniques that do not explicitly consider the topology of the palm vascular system, but they are pioneers in palm vein imaging. In~\cite{icprs2021style}, it is presented a GAN-based approach to simulate the texture of real images. The proposed model was trained with real images from the CASIA database~\cite{CASIA} to produce ROI samples of palm vein images. As an alternative, the second work in~\cite{icprs2021nature} aimed to reproduce the venous plexus of the palm by establishing an analogy with the patterns formed by the \textit{Physarum polycephalum}~\cite{A12}. The implementation is based on the algorithm developed in~\cite{liu2017new} in order to obtain vein patterns. Later, the method fuses the generated vascular pattern with a palmprint image to add details of the palm texture, obtaining a randomly-created sample of palm ROI. By using the proposed approaches, the authors created the largest palm vein databases of the state-of-the-art, as shown in Table~\ref{t:datasets}. Both synthetic datasets were validated through qualitative and quantitative experiments against the most relevant real databases. Taking into account our good previous results in these papers, in Section~\ref{sec:results}, we formalize a general flowchart for generating synthetic palm vein databases. 

{Although synthetic databases allow the evaluation of large-scale biometric recognition algorithms, to our knowledge, the vascular pattern images mentioned above have been used little for these purposes. Table~\ref{t:recognition-synthetic} summarizes of state-of-the-art approaches applied to biometric recognition using synthetic images of vascular patterns. While synthetic databases allow the evaluation of large-scale biometric recognition algorithms, to the best of our knowledge, synthetic vascular pattern images have been underutilized for these purposes. Actually, those investigations that cite synthetic databases either use them to improve synthetic databases or as part of state-of-the-art reviews~\cite{yang2020gan,Krishnan2020}. Focusing on hand veins (back of hand veins, palm veins, and finger veins), probably the low interest in the evaluation of biometric recognition algorithms is because the generation of synthetic vein images has many challenges that prevent an adequate estimation of parameters for their respective simulation. Among the challenges, it is highlighted that the real images present low quality, and the vascular patterns are not observable in the visible spectrum, requiring special devices for their visualization and capture, as shown in Section~\ref{ss:acquisition}.}

\begin{table}[ht!] %
\centering
\caption{{Summary of state-of-the-art approaches using synthetic vascular pattern databases for biometric recognition.}}\label{t:recognition-synthetic}
\scalebox{0.95}{
\footnotesize
\begin{tabular}{m{4cm}m{3cm}m{3.8cm}m{4.6cm}}
\toprule
\textbf{Reference model} & \textbf{Database} & \textbf{Biometric task} & \parbox{4.5cm}{\textbf{Performance\\ results}} \\
\midrule
Hillerstr$\ddot{\text{o}}$m~\etal~\cite{hillerstrom}, matching binary pattern & Database proposed by the authors & Finger vein verification & EER 1.63\%\\ \hline
Yang~\etal~\cite{yang2020gan}, ResNet34-based model & Database proposed by the authors & Finger vein verification & EER 0.24\%\\ \hline
Zhang~\etal~\cite{zhang2019}, FCGAN-CNN & Database proposed by the authors & Finger vein recognition and verification  & Accuracy 99.67\%, EER 0.52\%  \ \\ \hline
Das~\etal~\cite{das2017}, Support Vector Machine (SVMs) & Database proposed by the authors & Sclera pattern recognition and verification & Identification Accuracy 48\%, Verification Accuracy 75\% \\ \hline
\multirow{16}{4cm}{Hern{\'a}ndez~\etal~\cite{hernandez2021large}, AlexNet-based model and ResNet32-based model}  & \multirow{8}{3cm}{Synthetic--sPVDB~\cite{icprs2021style}} & \multirow{8}{*}{Palm vein recognition} & \textbf{--}~AlexNet-based model: \\
 &  & & \hspace{5mm}PolyU 96.00\% \\ 
 &  & & \hspace{5mm}PUT 93.00\% \\
 & & & \hspace{5mm}VERA 94.09\%\\
 & & & \textbf{--}~ResNet32-based model:\\
 &  & & \hspace{5mm}PolyU 99.90\% \\ 
 &  & & \hspace{5mm}PUT 99.00\% \\
 & & & \hspace{5mm}VERA 96.82\%\\
\cline{2-4}
 & \multirow{8}{3cm}{NS-PVDB~\cite{icprs2021nature}} & \multirow{8}{*}{Palm vein recognition} & \textbf{--}~AlexNet-based model: \\
 &  & & \hspace{5mm}PolyU 98.10\% \\ 
 &  & & \hspace{5mm}PUT 93.00\% \\
 & & & \hspace{5mm}VERA 94.09\%\\
 & & & \textbf{--}~ResNet32-based model:\\
 &  & & \hspace{5mm}PolyU 99.30\% \\ 
 &  & & \hspace{5mm}PUT 98.00\% \\
 & & & \hspace{5mm}VERA 95.45\%\\
\bottomrule
\end{tabular}}
\end{table}

{Hillerstr$\ddot{\text{o}}$m~\etal~\cite{hillerstrom} evaluate the performance of their synthetic finger vein database through an LBP-based model that retrieves and matches local binary patterns from previously enhanced synthetic images, which are obtained by histogram equalization. They then use the Chi-squared distance to calculate the matching scores. Estimate the matching accuracy using 225,000 genuine scores and 1,249,750,000 impostor scores from their synthetic database, achieving an EER of 1.63\%. These obtained results suggest the applicability of synthesized finger-vein images for large-scale biometrics applications.}

{Yang~\etal~\cite{yang2020gan} propose a model based on ResNet-34 and considers an experimental group and a control group. In the first group, the authors perform a pre-training using their synthetic finger vein dataset, followed by fine-tuning on the MMCBNU\_6000~\cite{MMCBNU} dataset. In the second group, training and testing are performed directly from the MMCBNU\_6000 database.  The performance verification results in the first group showed an EER of 0.24\%, while in the second group, the EER was 0.61\%. The above shows that recognition performance can be improved using synthetic images compared to direct training on MMCBNU\_6000.}

{Zhang~\etal~\cite{zhang2019} proposed a methodology for finger-vein image augmentation based on GAN networks called FCGAN. For biometric recognition, they a CNN trained with a two-stage progressive augmentation approach. 1) They perform image augmentation with the classical method (using translation in a different direction) and randomly distribute the images equally into ten mutually exclusive groups. Then, each group of finger vein images is fed into the CNN regularly to obtain an optimal result $\mathbb{M}^{(o)}$, as an achieving first-rank classification result. 2) $\mathbb{M}^{(o)}$ is trained sequentially as in the previous stage but using samples augmented with the FCGAN scheme. The authors compare their results with the SAGAN~\cite{zhang2019self}  and DCGAN~\cite{radford2015unsupervised}  methods of image magnification. The classification accuracy for SAGAN and DCGAN was 98.71\% and 98.74\%, respectively, while with the FCGAN scheme, they obtained an accuracy of 99.67\%. As for the false-positive and true-positive rates, the EER for the SAGAN and DCGAN methods were 0.93\% and 0.91\%, respectively, while with the FCGAN scheme, they obtained an EER of 0.52\%. The above demonstrates that the FCGAN augmentation method can generate higher quality and more diverse finger vein images, which improves the training performance of CNN.}

{Das~\etal~\cite{das2017} construct a synthetic database of eye sclera and use support vector machines as the biometric recognition algorithm. Here the authors use five synthetic images of the sclera of the eyes for training and five real images from the UBIRIS database~\cite{ proencca2005ubiris} for testing. Estimate the matching accuracy using 10*5 for false rejection rate (FRR) and 10*9*5 scores for false acceptance rate (FAR), thus achieving an accuracy of 48\% for identification and 75\% for verification. From the results obtained in this research, the synthetic image generation method was not adequate.}

{Finally, Hern{\'a}ndez~\etal~\cite{hernandez2021large} evaluated two end-to-end CNN architectures (based on AlexNet and ResNet32) on the Synthetic-sPVDB~\cite{icprs2021style} and NS-PVDB~\cite{icprs2021nature} datasets for biometric recognition. Both models were trained on the two synthetic databases and validated by a cross-testing scheme on the most representative real databases of the state-of-the-art. Experimental results are higher than 93\% for the AlexNet-based model trained on both synthetic datasets. In the case of the ResNet32-based model, reported results are comparable with those reported in the literature on the public databases.}

{Although using synthetic vein pattern databases in evaluating biometric recognition algorithms is still limited, state-of-the-art results described above show the feasibility of their use in vascular-based biometrics.}

\subsection{Generation of Vein Patterns Related to Biological Networks}\label{ss:nature}

The formation of venous structures with a perspective oriented to biological transport is relevant in the biometric context because they reproduce patterns and predict measurable quantities. Biological transport networks are part of the critical infrastructure required to distribute resources quickly and efficiently throughout the body. They include the vascular systems of plants, animals, plasmodial veins of slime molds, among others. Vascular networks play a key role in the overall organization of the body. Furthermore, there is evidence that vessel size and network architecture follow fourth power allometric scaling laws that relate the fractal branching pattern of the network to body size and metabolic rate according to metabolic scaling theory  (MST)~\cite{A1}, or reflect the design of optimal transport networks~\cite{A2,A3,A4}.

A venous network is composed of tubular elements with different thicknesses and lengths organized as a hierarchical branching tree. The structure obtained can be open or closed, with a significant number of additional loops to improve performance under fluctuating loads or in the presence of damage~\cite{A5,A6}. Blood vessels have geometric characteristics that impact local blood flow, although their general flow behavior and scale relationships exhibit other networks' properties.

\begin{figure*}[!ht]
    \centering
    \includegraphics[width=0.85\textwidth]{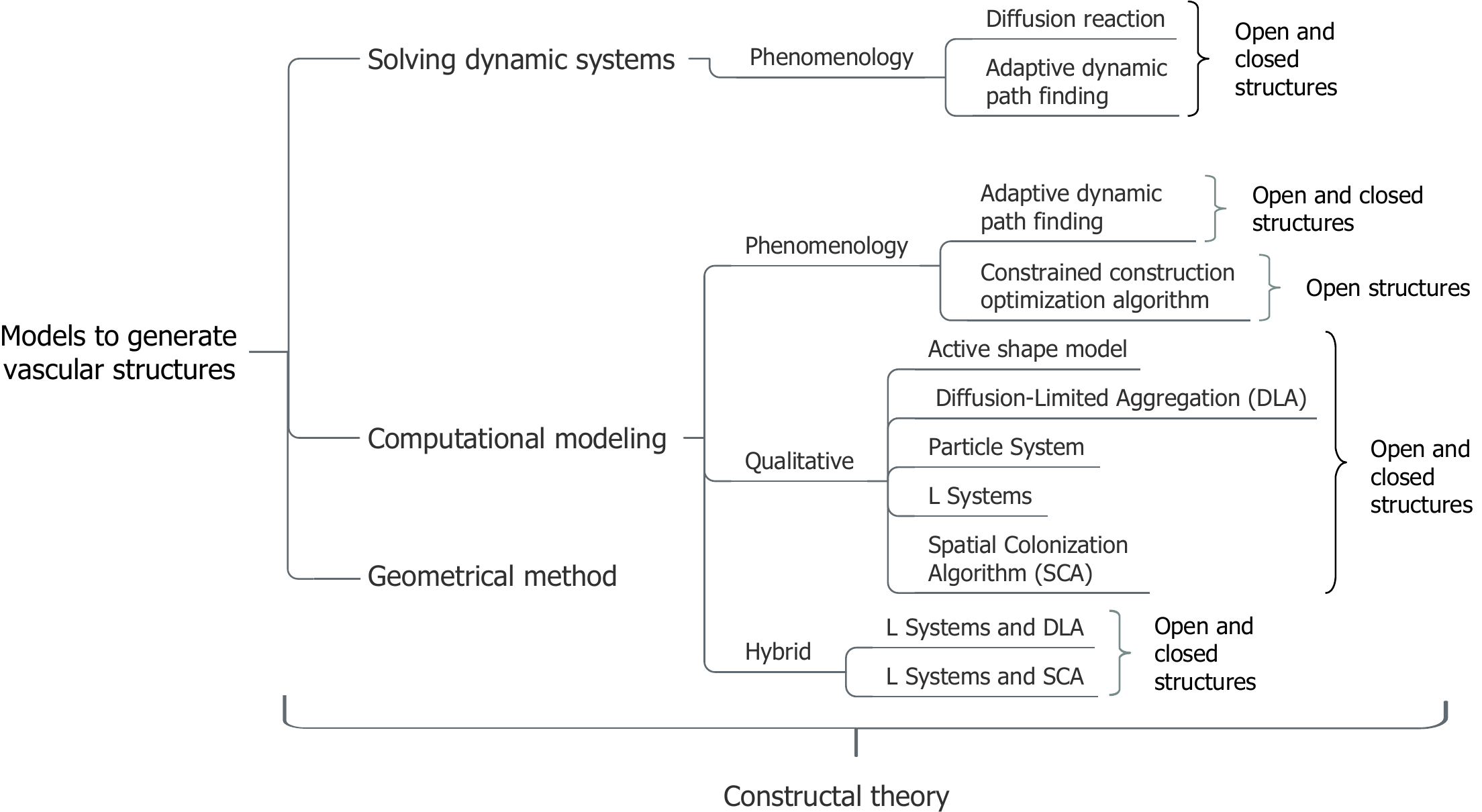}
    \caption{Classification of models used in the generation of biological vascular structures.}\label{mapa}
\end{figure*}

There are a wide variety of models that allow the representation of open and closed vascular structures. This section presents an updated review of models, and we have classified them into three categories: (i) solving dynamical systems, (ii) computational modeling, and (iii) geometrical methods. In Figure~\ref{mapa}, the models have been classified according to these categories and taking into account the parameters associated with the models, which can be phenomenological, qualitative, or hybrid (combination of models). These models can be considered part of a constructive theory, which maintains that every system is destined to remain imperfect. However, nature distributes them optimally from the global optimization of local constraints~\cite{chen2012progress}.  

The networks generated by the numerical resolution of dynamical systems are based on the assumption that the network evolution can be represented by partial differential equations that model the growth phenomenology~\cite{scianna}. Computational modeling consists of producing discrete events through a series of dynamic attributes (phenomenological or qualitative) defined locally on a dataset; this allows the network to evolve over time with a minimum of a priori assumptions~\cite{runions2005}. Likewise, we have classified the models considering their applications and parameters in Table~\ref{braching_model}.

\begin{table}[ht!]
\centering
\caption{Summary of the state-of-the-art models for the generation of biological vascular structures.}\label{braching_model}
\scalebox{0.9}{%
\footnotesize
\begin{tabular}{m{3.3cm} m{2.2cm} m{5cm} m{5.5cm}}
\toprule
\textbf{Model} & \textbf{Type}  &  \textbf{Parameters} & \textbf{Application}  \\ \midrule
Reaction-diffusion systems~\cite{turing} & Solving dynamic systems & Diffusion rates and reaction parameters (nonlinear functions) & Cell growth~\cite{turing}, plants venation~\cite{Meinhardt}, and wing vein patterns~\cite{kondo}  \\ \hline

Adaptive dynamic pathfinding~\cite{tero2006physarum,jones2010characteristics} & Solving dynamic systems and computational modeling & Concentration and nutrient gradients & Pattern generation formed by the foraging of some multicellular organisms (\ie~\textit{Physarum polycephalum})~\cite{tero2006physarum,jones2010characteristics} \\ \hline

Constrained constructive optimisation~\cite{de2015} & Computational modeling  & Intravascular volume and branching angles (cost function) & Complex arterial trees~\cite{bui2010development,talou2021}, human
cerebrovasculature~\cite{ii2020multiscale} \\ \hline

Active shape model~\cite{cootes1995active} & Computational modeling & Reference points of real images  & Retinal vascular network~\cite{bonaldi2016automatic} \\ \hline

Diffusion-limited aggregation~\cite{witten1981} & Computational modeling & Density correlations  in terms of the distance  separating the two sites & Branching structures (arteries/veins)~\cite{bourke2006}, river networks~\cite{pelletier2000}, simulation of natural texture~\cite{roberts2001sticky}  \\ \hline

Particle system~\cite{rodkaew2002algorithm} & Computational modeling & Particle distribution and motion & Leaf structures~\cite{rodkaew2002algorithm} and topology optimization of heat conduction systems~\cite{lohan2017topology} \\ \hline

Lindenmayer system (L-system)~\cite{prusinkiewicz} & Computational modeling & Branching angles, elongation rates and vigor of branches & Virtual plants~\cite{prusinkiewicz}, retinal  vasculature~\cite{aghamirmohammadali2018} and blood vessel~\cite{galarreta2013three} \\ \hline

 Space colonization algorithm~\cite{runions2007} & Computational modeling &  Attraction points, radius of influence and kill distance & Branching structure of trees~\cite{runions2007}, leaf venation~\cite{runions2005}, finger vein images~\cite{hillerstrom} and tracheal gills of mayfly larvae~\cite{ruiz2020space} \\ \hline

Hybrid models~\cite{jin2009improved,salcedo2019hybrid} & Computational modeling & Parameters associated with the models: L-system--Diffusion-limited aggregation and L-system--Space colonization algorithm & Leaf venation patterns~\cite{jin2009improved} and complex structures~\cite{salcedo2019hybrid}. \\ \hline

B-spline based vascular structures~\cite{li2007mathematical,castro2020visual} & Geometrical method & Control points or vector of nodes & Flexible vascular representation~\cite{li2007mathematical}, Retinal vascular network~\cite{castro2020visual} \\ 

\bottomrule
\end{tabular}%
}
\end{table}

Among the models for solving dynamic systems to generate vascular structures are those based on reaction-diffusion methods~\cite{turing}. These models produce theoretical open and closed vascular structures from phenomenological principles with realistic parameter values~\cite{anderson1998continuous}. Based on this approach, vascular networks are the result of two mechanisms: (i) vasculogenesis, which is the formation of new blood vessels from a population of endothelial cells, and (ii) angiogenesis, which is the formation of new blood vessels from pre-existing vessels. In the case of vasculogenesis, it is assumed that the mechanism driving vascular network formation is due to cell traction forces that provide local chemical signals that attract cells to initiate venous patterning~\cite{Gamba2003}.

Applying reaction-diffusion models present some weakness. The vasculature contains a range of spatial scales, which cannot be controlled respecting allometric scale laws~\cite{A1}. In addition, a highly interconnected network is assumed that vessel pressures at nearby spatial locations are correlated~\cite{shipley2020hybrid}. We emphasize that modeling through dynamical systems allows the generation of closed structures. Therefore, there is a particular interest in this type of model producing vascular networks assuming anisotropic structures such as those formed in the skin~\cite{Lanza2006}. The method assumes that external chemicals in the environment attract/repel and exert drift forces that direct the formation of vascular networks to specific sites. The general model is defined by the system of partial differential equations as follows,

\begin{equation}
\left.
\begin{array}{ccc}
  \partial_{t} \eta &=& -\nabla\cdot(\eta\mathbf{v}) \\
  \partial_{t}\mathbf{v}  &=&\mathbf{v}\cdot\nabla\mathbf{v}+\boldsymbol\beta\cdot\nabla\mathbf{c}-\gamma\mathbf{v}-\nabla \psi(\eta)\\
\partial_{t}\mathbf{c}&=& \mathbf{D}\Delta\mathbf{c}+f(\mathbf{c},\eta)
\end{array}
\right\}
\end{equation}

where $\eta$ is the endothelial cell population, $\bf{v}$ defines the flow direction, $\boldsymbol\beta$ is the intensity of the cellular response, $\gamma$ is the drag factor with respect to the substrate, $\nabla \psi(\eta)$ measures the response of the set of cells to deformations, $\bf{c}$ is the vector of chemoattractants-chemorepellants, $\mathbf{D}$ is a diagonal matrix containing the diffusion coefficients, and $f(\mathbf{c},\eta)$ is the function that induces the reaction between $\eta$ and $\mathbf{c}$.

As part of reaction-diffusion systems, Meinhardt~\cite{Meinhardt2008} introduced a new model based on activator-inhibitor and activator-substrate chemical gradients to generate network-shaped structures that allow the water supply, oxygen, and information on a tissue. The orientation results from local signals that spread as the network grows, and the cells respond by differentiating into members of the branch system. Differentiated cells are supposed to eliminate growth signal-generating tropic substances, making the system sensitive to minute differences in concentration. The formation of branches occurs in two ways: lateral branching if new growth signals are activated along the branches, or a dichotomous branching if the growth signal is divided at the ends of the branches. Among the applications that have been given to the model are the generation of blood networks, insect tracheas, axons, or leaf veins~\cite{Meinhardt, Meinhardt2008, jonsson2010modeling}.

Moreover, the systems based on activator-inhibitor and activator-substrate principles have shown how interactions at the molecular level can lead to morphogenesis and differentiation for angiogenesis. Dorraki~\etal~\cite{Dorraki2020angiogenic} review the dynamics of angiogenesis to study tumor growth, where they show models of continuous focus with other parameters such as components of the extracellular matrix, oxygen, migration, and proliferation of endothelial cells. The models considered in~\cite{Dorraki2020angiogenic} although describing the dynamics better, are not intended to generate branched structures.

Pathfinder-based models with adaptive dynamics are representative of both resolution of dynamical systems and computational modeling (see Figure~\ref{mapa}). For generating a network, the parameters used are characteristic of phenomenology and allow the construction of open and closed structures. Such models are inspired by the conduct of organisms whose survival depends on optimally seeking pathways to nutrient sources. These kinds of adaptive dynamics have also become used in the construction of heuristic algorithms in optimization processes~\cite{serrano2019meta,beheshti2013review}. In this sense, an algorithm based on \textit{Physarum polycephalum} has allowed the creation of vein networks similar to those found in living beings~\cite{tero2006physarum,jones2010characteristics}. 

The growth of Physarum has been modeled through the resolution of dynamical systems assuming Jean-Baptiste Lamarck's first law, which establishes the development of animal organs based on their utility~\cite{burkhardt2013lamarck}. Lamarck's first law could be interpreted as the vessels' diameters increasing depending on the flow rate~\cite{Akita2017study}. The network begins with sources and sinks located in an arbitrary mesh generated by a Delaunay triangulation. The tubular thickness adaptation is a function of the conductivity at the node $(i,j)$ that changes in time according to the expression

\begin{equation}\label{eq:physarum}
  \frac{dD_{ij}}{dt}=f\left(|Q_{ij}|\right)-D_{ij}
\end{equation}

where $D_{ij}$ is the fluid conductivity and $f$ is a continuous monotonic increasing function that depends on the flow rate $Q_{ij}$ and satisfies $f (0) = 0$. In addition, $f$ represents a coarsening factor that increases with the flow, and $-D_{ij}$ is a thinning factor. The coarsening component dominates for a channel with a larger flow, and the thinning effect dominates for small flows at each time step. The complete time evolution of the network is calculated iteratively with the updated conductivities.

On the other hand, computational modeling of pathfinding with adaptive dynamics employs an agent-based layer and a path layer that affect each other. Thus, agents deposit a trace on the path map, while those particles detect values from the map to determine aspects of their movement~\cite{jones2010characteristics}. Each agent has a heading angle, a location, and three sensors (front left, front, front right). The sensor readings affect the particle's heading, causing it to turn left, turn right, or stay in the same direction. The paths undergo a diffusion and decay process at each time step.

Also within the computational modeling context are constrained constructive optimization algorithms, which use parameters that depend on phenomenological features and allow the generation of open structures. This approach produces complex dichotomously branched arterial tree models that have been shown to reproduce key structural features of coronary arterial trees~\cite{schreiner1994}. Network generation is sequential and driven by the constrained minimization of a cost function designed to minimize the energy expenditure of blood supply to surrounding tissues~\cite{talou2021}. The construction of a model tree is achieved by the stepwise addition of individual segments that simulates natural growth. At each step, this growth is guided by an optimization objective and by ensuring the growing tree meets geometric conditions between vessel diameters at a bifurcation following Murray's law, as well as a specific pressure or flow at the tree terminations~\cite{schreiner1997}.  A set of pseudorandom numbers is used to determine the sites of new segments, which become the main branches of the fully developed tree in the early stages of development.  Therefore, the topography of these main branches is highly dependent on the set of pseudorandom numbers used~\cite{de2015}. 

Constructive constrained optimization algorithms address different types of studies involving the influence of anatomical variability, bifurcation asymmetries, fractal properties, and stepwise growth~\cite{talou2021}. Also, some variants have been proposed to recreate more complex vascular networks in hollow organs or to accelerate the construction of such networks using heuristic optimization algorithms~\cite{ii2020multiscale}. Particularly, Bui~\etal~\cite{bui2010development} used a constrained construction optimization algorithm to produce cerebral vasculature with a preferential distribution of large arteries in the cerebral cortex. Besides, Georg~\etal~\cite{georg2010global} simulate a hepatic venous tree generated in stages following fractal behavior.

An alternative way to produce vein networks is using active shape models, which are part of computational modeling and provide a statistical representation of shapes from real image landmarks, allowing both open and closed vascular structures to be generated~\cite{cootes1995active}. Training samples are aligned in a common coordinate frame, and deviations from the mean shape are analyzed. Each training shape is represented as a fixed number of $n$ landmarks placed along an equally spaced grid. These reference points form a vector, which is subsequently reduced by principal component analysis (PCA), assuming that the most relevant features are those with the highest variance. Such models allow a variety of structures to be generated since they use reference images. Particularly, in~\cite{bonaldi2016automatic} they use an active shape model to generate synthetic images of the retina. The disadvantage of the method is that good-quality images are required for segmenting the vascular networks to reproduce the desired structure. While allowing the generation of vascular structures, the previous model is not explicitly focused on this but rather on reproducing variations of any shape. In contrast to this approach, the following will describe computational models specifically focused on vascular structures.

Diffusion limited by aggregation (DLA) is a process attributed to Forrest and Written~\cite{witten1981}. This process involves subjecting a set of particles to a random walk so that they subsequently agglomerate around a particle called the seed. The DLA starts with the seed at a fixed position within a bounded region. The particles are then randomly released from the edge and follow a Brownian motion until reaching the seed; highly branched or solid structures can be achieved by introducing a sticking probability~\cite{bourke2006, sidoravicius2019}. Other variants of this approach have also emerged, such as Cluster-Cluster Aggregation~\cite{lin2014universality}, which starts with a system of particles that diffuse in space and form clusters when they meet.  These, in turn, can interact with other clusters or particles to form even larger aggregates. In~\cite{roberts2001sticky}, the authors propose the Random Drop Ballistic Aggregation algorithm. Instead of using a Brownian motion to form the aggregate, they define a radius and direction on the particles to choose the closest adhesive position of the cluster. Their method uses a kernel that can change the adhesion conditions, limiting the growth according to the clusters and inhibiting the formation of long tendrils.

As in the continuous models, the disadvantage of the DLA algorithm is that it is not possible to control the allometric scales. Despite this, network growth by DLA tends to occur at the extremes of the structure, and particles are less likely to reach the inner parts. This feature operates across scales and gives rise to a fractal nature, which can be associated with physical and chemical properties. This model is used to simulate growth patterns that appear in some organic and non-organic phenomena such as crystal growth, the trajectory described by lightning, vascular systems, among others~\cite{bourke2006}.  Particularly, in~\cite{pelletier2000}  show that the structures of river and leaf vein networks are statistically similar and can be simulated on a small scale using aggregation-limited diffusion.

Lindenmayer systems or L-systems allow studying various aspects of vascular pattern formation following a fractal behavior~\cite{prusinkiewicz}. The branching information and changes in topology are realized using a mathematical construct called a rewrite or production rule. A rule is written as predecessor--successor; the successor in the resulting string iteratively replaces the predecessor. On the other hand, to account for external factors such as length and angle dimensions that may vary at each rewrite step, a perimetric L-system is defined~\cite{prusinkiewicz}. Among the applications of L-systems are the generation of 3D synthetic blood vessels by adding stochastic parameters~\cite{galarreta2013three} and the generation of retinal venous structures, which is relevant in diagnosing and treating diseases affecting the eyes. The retinal vasculature is realized by an inverse L-system problem, in which the rules must be obtained from images representing the tree-like structure~\cite{aghamirmohammadali2018}.
 
The particle system is an algorithm for generating leaf vein patterns for realistic synthetic images~\cite{rodkaew2002algorithm}. This algorithm initially distributes a set of particles over a leaf-shaped region. The particles move towards a sink located at the base. In the process of displacement, they attract each other and merge when they reach a distance threshold. The venation pattern is the result of the trajectories taken by the particles. Some of the patterns generated develop primary and secondary veins that are typical of open venation structures. This algorithm has also been used in the topological optimization of structures for heat conduction~\cite{lohan2017topology}.

The spatial colonization algorithm is a model for generating highly irregular leaf venation patterns~\cite{runions2005}. The model works by iteratively extending the partially formed veins to points considered sources of a vein-inducing signal. Sources are dynamically added as the leaf grows and removed as veins approach them. Also, the algorithm generates morphological properties, such as each branch's length, diameter, and angles, through Murray's law. The branching structures have a hierarchical order, with the first-order veins being the largest in diameter. Higher-order veins may have a free end, producing an open veins pattern or anastomose, forming loops characteristic of a closed pattern. The parameters taken into account in the algorithm are the attraction distance, the elimination distance, and the length of the segments. The spatial colonization algorithm has also allowed the generation of other types of structures. In~\cite{hillerstrom}, they use the algorithm for the generation of synthetic images of finger veins, and in~\cite{ruiz2020space}  simulate the branching patterns of mayfly gills to investigate the global mechanism driving their generation and morphological evolution.

The hybrid models combine two or more models to improve both the performance and the quality of the vein network structure. In~\cite{jin2009improved} they propose a method that combines spatial optimization algorithm with L-systems to accelerate the growth of the leaf vein network. Their results show that the properties of the network can be improved in a minimum time. In~\cite{salcedo2019hybrid} combine L-systems with DLA to model novel growth structures. Particularly, L-systems are used to guide DLA simulations~\cite{salcedo2019hybrid}.

Finally, B-spline-based mathematical models are segmental interpolators generating smooth curves and using control points to form realistic vascular structures~\cite{li2007mathematical,castro2020visual}. The control points are defined at main points in the network, such as bifurcations or terminal points. Different B-spline types have been used for the elaboration of vascular networks. In~\cite{li2007mathematical}, the authors use non-uniform rational B-spline (NURBS) to reproduce vascular network models based on physiological or image-based parameters. NURBS allows to adjustment of the shape of a curve through a weight parameter. On the other hand, in~\cite{castro2020visual} introduce the Catmull Rom spline (a particular case of NURBS, when the weight parameter is equal to zero) to simulate the Retinal vascular network, determining the control points from the temporal arcade. Although B-spline allows the generation of realistic vascular networks, it is not a model for generating vein networks. Instead, it is a technique that allows efficient point along with the characteristic that the resulting curve is differentiable at all points, thus providing images with a high degree of realism.

\section{Evaluation of General-Purpose Approaches for Generating Synthetic Palm Vein Images}\label{sec:results}

\begin{figure}[!ht]
    \centering
    \includegraphics[width=0.7\textwidth]{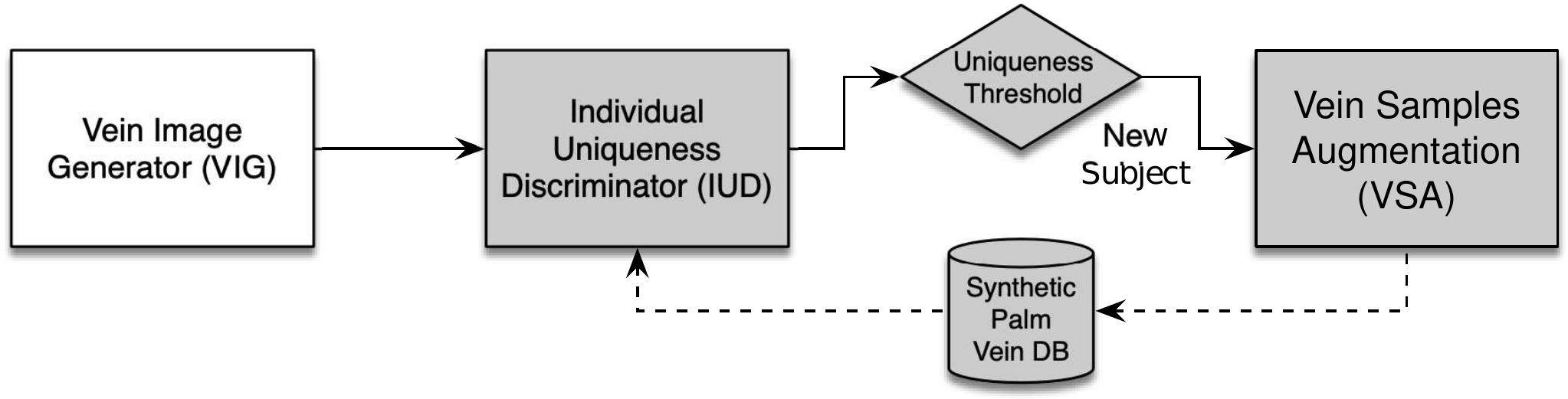}
    \caption{Overview of the flowchart of studied approaches for generating synthetic palm vein images and creating a synthetic database. The processes highlighted in gray color are common to both approaches, whereas the VIG process varies depending on the generation method implemented.}\label{f:model}
\end{figure}

In the following, we present and evaluate two different approaches for the generation of synthetic palm vein images, which were firstly introduced in our conference papers~\cite{icprs2021style, icprs2021nature}. Based on our previous results and the review of the state-of-the-art, we formalize a general flowchart for the creation of a synthetic database, as shown in Figure~\ref{f:model}. The proposed methods are focused on generating region-of-interest (ROI) samples of palm vein images since usually SOTA methods for palm vein recognition use the ROI from a human hand.

The purpose of this section is not to provide a large-scale database of palm vein images per se, and in any case, do not replace datasets of real-world images. Thus, it aims to explore options from the state-of-the-art to generate palm vein samples, both computational and mathematical methods. Hence, the created synthetic datasets can be alternatives to quickly estimate the efficiency, scalability, and general performance of newly or formerly designed approaches for palm vein recognition on a larger dataset than the currently publicly available ones. 

The proposed flowchart comprises three main procedures; whereas the first one varies depending on the generation method implemented, the last two processes are common to both approaches. Each process is described as follows:

\begin{enumerate}
    \item \textbf{Vein Image Generator (VIG)}: The VIG procedure aims to generate random synthetic images that simulate both the vein patterns (\ie~dark gray patterns of vascular structures) as well as visual characteristics (\ie~texture, contrast, and illumination) of palm vein images. As was analyzed in Section~\ref{s:review}, it is important to notice that there are different methods to produce synthetic vascular structures. Thus, the presented model can be implemented using a variety of approaches by modifying the generator in correspondence with the other parameters involved in acquiring real images of palm vein patterns. Therefore, we evaluate two different general-purpose methods to implement the VIG procedure. We firstly analyze a novel specific domain application for the style-based GAN architecture (StyleGAN)~\cite{karras2019style} for generating synthetic palm vein images, which is presented in Section~\ref{ss:style-synthetic}. Under the foundation that the acquisition process produces images with very particular characteristics in terms of texture, illumination, and contrast, this approach aims to generate synthetic palm vein images of unique individuals, simulating the same style of the real images. Later, in Section~\ref{ss:nature-synthetic}, we examine a nature-based approach for simulating the palm vein plexus, which is inspired in the natural transport network by the growth of the \textit{Physarum polycephalum} that forms path patterns similar to vascular structures.
    
    \item \textbf{Individual Uniqueness Discriminator (IUD)}: The previous procedure randomly generates synthetic palm vein images, satisfying the uniqueness requirement to be part of a biometric database. Therefore, the IUD procedure compares each newly produced image against all samples in the synthetic database to ensure the uniqueness requirement. Since the process of creating the synthetic database is incremental, we adopt a matching procedure based on a uniqueness threshold by computing the similarity score proposed in~\cite{hernandez2019individuals}. In accordance with the experiments of~\cite{hernandez2019individuals}, we fix the uniqueness threshold to 0.1 as a good separability criterion of genuine and impostor samples. Hence, the generated image is discarded if the computed similarity score with any instance in the database is higher than the uniqueness threshold. Otherwise, if the threshold is not exceeded for any of the samples in the database, the generated image satisfies the uniqueness criteria and is incorporated into the database as a new subject. Because an exhaustive search must perform the similarity matching process in the database, we implemented a multi-thread algorithm~\cite{hernandez2019individuals} to reduce processing time.
    
    \item \textbf{Vein Samples Augmentation (VSA)}: Since a biometric database should contain more than one sample per individual (\eg~gallery and probe samples), in order to increase the number of samples per subject, we perform a sample augmentation process for each new unique image generated by the IUD procedure. Thus, the VSA procedure aims to obtain different samples for each subject by applying random transformations on the image to simulate natural variations of a real-world contactless acquisition process of palm vein images. For this purpose, two kinds of transformations are randomly applied over the primary sample: texture/illumination variations (\ie~contrast, lighting, and blurring), to simulate the effect of NIR illumination, and affine transformations of the ROI (\ie~translation, rotation, cropping, and shear), to replicate changes of the hand position on the acquisition device. Finally, we obtain six samples per each new subject in the database.
\end{enumerate}

\subsection{Generation of Style-based Synthetic Palm Vein Images}\label{ss:style-synthetic}

As mentioned in Section~\ref{ss:RedesBiometria}, different approaches have adopted GAN variants for sample augmentation purposes in finger-vein recognition. Besides, GAN-based methods have been proposed for generating iris~\cite{kohli2017synthetic, yadav2019synthesizing} and face images~\cite{yang2020faces, varkarakis2020refaces}. Thus, taking into account their previous proposals, we study the effectiveness of the StyleGAN architecture~\cite{karras2019style} to implement the VIG procedure, aiming to simulate the visual characteristics of palm vein images in terms of texture, illumination, and contrast. Among GAN variants, the StyleGAN2 model~\cite{karras2020analyzing} has demonstrated to achieve better results in terms of image quality, images variability, and efficiency, improving the former architecture and generating more realistic images. Hence, we firstly implemented a StyleGAN-based vein image generator (Style-VIG) aiming to create random palm vein samples and reproduce the \textit{style} of real palm vein images with a high level of realism.

\begin{figure}[!ht]
    \centering
    \includegraphics[width=0.8\textwidth]{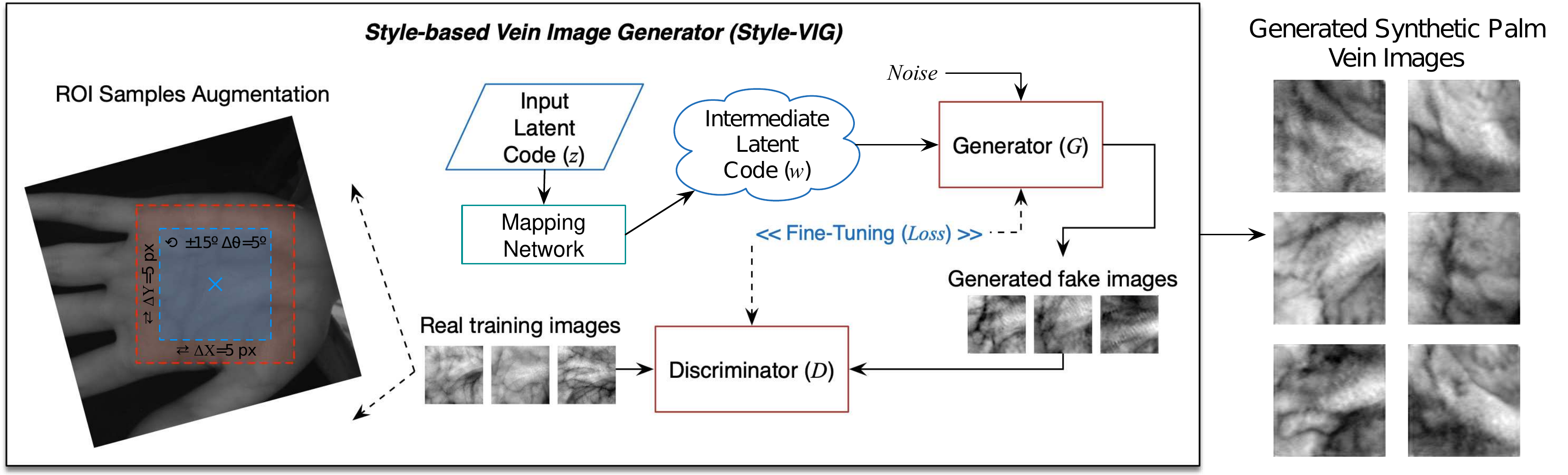}
    \caption{General scheme of the StyleGAN-based Vein Image Generator (Style-VIG), which is based on the StyleGAN2 model.}\label{f:style-vig}
\end{figure}

Figure~\ref{f:style-vig} depicts the GAN-based approach implemented by the Style-VIG process. Firstly, the StyleGAN model transforms the Input Latent Code ($z$) by mapping it into an Intermediate Latent Code ($w$), which encodes the style of the real images. Thus, the Generator ($G$) synthesizes new samples from the mapped coded ($w$) by adding random noise to facilitate the stochastic variation. The StyleGAN2 architecture improves the original StyleGAN model by applying the bias and noise operations on normalized data used in the generator, which helps to remove normalization artifacts from generated images. On the other hand, the Discriminator ($D$) learns to distinguish real and fake images by training a binary classifier from real training images. Thereby, both networks are trained collaboratively meanwhile fine-tuning each other by minimizing the GAN loss function.

Due to public datasets of real images have a limited number of samples, a challenge to face is the lacking of training data. Hence, different from other approaches that apply image augmentation from the ROI samples during the training process, we implemented a ROI sample augmentation procedure during the pre-processing stage. Thus, we obtain 87x ROI samples from each raw image by translating and rotating a central ROI (blue square) within the entire palm (red square). The central ROI is centered on the palm region, which its side is determined by the middle points between the index and middle finger, and the ring and little finger, respectively. To augment the number of ROI samples, the central ROI is repeatedly shifted in both axis with $\Delta(X,Y)=5px$, and also it is rotated by $\Delta \theta=5^{\circ}$ up to $\pm 15^{\circ}$, both clockwise and counter-clockwise.

In our experiments, we trained the StyleGAN2 network~\cite{karras2020analyzing} on the 850nm-band of CASIA dataset~\cite{CASIA}, which is commonly used for palm vein recognition. Hence, the training data was originally comprised of 1,200 real images, which were increased up to 104,400 training samples after performing the procedure for ROI samples augmentation. Our implementation was based on the official StyleGAN2 source-code from NVIDIA Research Projects\footnote{\url{https://github.com/NVlabs/stylegan2}}. We used the original configuration (config-e) of the architecture to train the StyleGAN2 model for 25,000 k-images. Finally, the training process reported a Frechet Inception Distance (FID) equal to 63.89. Whereas the training process took five days, the generation process takes on average 58.1ms ($\pm$ 259.69ms) to create a new synthetic sample.

Besides, as part of the present work, we extended the Synthetic-sPVDB dataset~\cite{icprs2021style}, reaching a total of 20,000 individuals and 120,000 synthetic samples. This second version has been possible thanks to the incremental generation process aiming to increase the number of individuals in the database. As is noticeable in Table~\ref{t:datasets}, compared to the largest real database Tongji~\cite{Tongji}, the proposed dataset is the largest of the state-of-the-art, being 33 and 10 times larger in the number of subjects and the total samples, respectively.

\subsection{Generation of Nature-based Synthetic Palm Vein Images}\label{ss:nature-synthetic}

As the second approach for implementing the VIG procedure, we developed a slightly more specific method based on biological transport networks. In this case, we aimed to examine a mathematical model for simulating the palm vein plexus and comparing the previous general-purpose computational model against a natural-inspired one. The idea behind this method is based on the principle that the vascular network adapts its structure in an optimal way to the hand's movement and muscles. Thus, we implemented a mathematical model based on pathfinding with adaptive dynamics, which have been used to simulate the growth of the \textit{Physarum polycephalum}~\cite{A12}. Particularly, the \textit{Physarum polycephalum} dynamically forms regular path patterns similar to the vein plexus network since its growth patterns establish optimal organizational structures with high efficiency.

\begin{figure*}[!ht]
    \centering
    \includegraphics[width=0.8\textwidth]{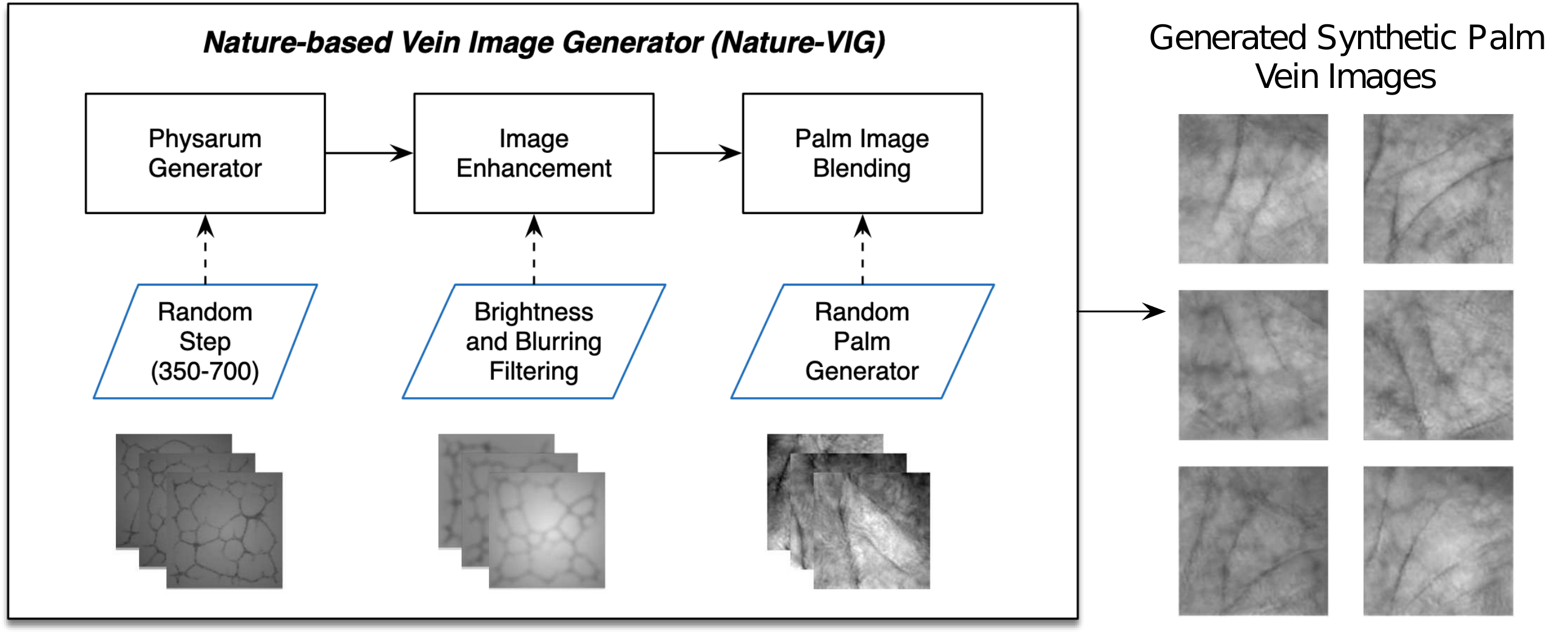}
    \caption{General scheme of the Nature-based Vein Image Generator (Nature-VIG), which is based on the growth of the \textit{Physarum polycephalum}.}\label{f:nature-vig}
\end{figure*}

In the literature, there are different mathematical approaches for modeling the growth of the Physarum~\cite{tero2007,gunji2008minimal,liu2017new}. Particularly in~\cite{liu2017new}, the authors propose a multi-agent model to simulate the growth of the network structure. The proposed model uses a simple process to produce the complex patterns of the Physarum, which is based on two requirements: optimal searching of nutrients within the maximum area and the loop formation to ensure a fault-tolerant transport network. Thus, since palm vein patterns should satisfy similar optimization processes to supply blood to the entire surface of the hand, we used the mathematical model of~\cite{liu2017new} to simulate an approximate structure of the palm vein patterns. An overview of the proposed nature-based VIG process is shown in Figure~\ref{f:nature-vig}, which is called Nature-VIG and is comprised of three sub-processes described following.

Firstly, the Physarum Generator is used to generate a vascular network by using the multi-agent model proposed in~\cite{liu2017new}. The multi-agent model is based on a self-organized system approach, which uses two types of agents and three transition rules to simulate the network's behaviors and interaction, respectively. The growth of Physarum is characterized by adaptive search and contraction behaviors considering global and local conditions of the environment. Initially, the agents randomly move within the square representing the palm ROI and then converge to the most optimal network to connect the food sources located in the midpoints of the top and bottom of the region, which analogously correspond to the proximal to the distal part of the palm. This random search process is performed iteratively, where the higher the number of iterations, the more different types of patterns are formed due to the contraction behavior. According to the experiments performed in~\cite{liu2017new}, for more than 200 iterations, the dynamic change of the agents tend to converge to a steady pattern by reducing the number of loops. Hence, in our model, we set this parameter by a random number from 350 to 700 iterations, which, jointly with the initial random movement of the agents, ensure the inter-class variations of the generated samples.

The second sub-process of the Nature-VIG procedure aims to enhance the vascular patterns produced by the Physarum Generator. We apply both brightness and blurring filters over the vascular image to reduce the discontinuities formed by the agents' imprint obtaining a continuous vascular structure and simulating the NIR illumination effects occurring in the acquisition of real images, respectively. Thus, the brightness variations are randomly applied, increasing the overall illumination between 1.2 and 1.8 as a factor of the original brightness. Afterward, we compute a sequence of extended box blurring filters over the obtained images, which uses a Gaussian kernel (with \textit{radius} $=3 \sim 5$)~\cite{blurringfiltering}.

Finally, to obtain the final palm vein image, we blend the previous enhanced image with a randomly generated palmprint image in order to add palmprint details and texture. For this purpose, we trained a StyleGAN2 model~\cite{karras2020analyzing} to randomly generate palmprint images based on the images of the CASIA database~\cite{CASIA}. For training the GAN model, we used ROI samples from the visible spectrum (WHT) of CASIA because they were collected with the same acquisition protocol of palm vein images in the dataset. Thus, to augment the number of training samples, we employ the same image augmentation process described for the Style-VIG process (Section~\ref{ss:style-synthetic}), obtaining 104,400 samples from 1,200 palmprint original samples. As well as for the Style-VIG process, our implementation of the StyleGAN2 model was based on the NVIDIA Labs' implementation. 
The model was trained for 25,000 k-images with a final FID equal to 77.28. Lastly, after having executed all the sub-processes of the Nature-VIG procedure, we get the final image of synthetic palm vein patterns on average time up to 179.32 ($\pm$ 37.29) seconds.

As for the previous generation method, we have increased the size of the NS-PVDB database obtained by the Nature-VIG procedure. Thus, we generated 96,000 synthetic samples corresponding to 16,000 different subjects. Since the nature-based simulation of vein patterns is more time-consuming than the GAN-based approach, it has not been possible to achieve the same size as for Synthetic-sPVDB; however, we will continue to increase the dataset's size.

\subsection{Performance Analysis of Generated Synthetic Images}

In~\cite{icprs2021style} and~\cite{icprs2021nature}, we performed both qualitative and quantitative evaluations to compare the similarity between real and synthetic datasets. For the qualitative assessment, we used the Gray-Level Co-occurrence Matrix (GLCM)~\cite{GLCM}, the Deep Image Structure and Texture Similarity (DISTS) index~\cite{DISTS}, and the Structural Similarity Index (SSIM)~\cite{SSIM}. Also, we compute quantitative metrics such as the Frechet Inception Distance (FID)~\cite{FID}, the Accuracy of a binary classifier to predict real and generated samples, and the $F_{1/8}$ and $F_8$ metrics proposed in~\cite{PRD}. In the evaluation, we compared the generated datasets against the most relevant public databases. The experimental results suggested that synthetic and real palm vein samples share visual features verifying the realistic characteristics of the generated datasets.

The following sections present an extension of the previous validation of the proposed synthetic datasets~\cite{icprs2021style,icprs2021nature}. We carry out new experiments aiming to deepen our understanding of the similarity between real and synthetic palm vein samples. First, we conduct a human inspection of random samples from the public database and the generated datasets to understand the visual perception for classifying both sets of images. Second, we analytically model the lower-order similarity of textures by using Bessel K forms~\cite{srivastava2002stochastic, srivastava2002universal}. Finally, the evaluated quantitative metrics are computed on the databases not considered in the conference papers~\cite{icprs2021style,icprs2021nature}. 

The experiments were performed on a dedicated server comprised of two Intel Xeon Gold 6140 CPUs (36 physical cores), 126GB of RAM, and four NVIDIA GeForce GTX 1080Ti GPUs. The models were implemented using Python 3.7 and OpenCV 4.1, while the implementation of StyleGAN required TensorFlow 1.15, CUDA 10.1 toolkit, and cuDNN 7.5. Additionally, the created synthetic datasets are available online at: \url{http://www.litrp.cl/repository.html}.

\subsubsection{Visual Evaluation by Human Inspection}\label{ss:visual_results}

Intending to examine the human perception of palm vein images produced by both generation methods studied, we developed a human inspection experiment to classify ROI images of palm veins into real or synthetic. For this purpose, we invited 40 people separated into two groups according to their experience related to image processing and vascular biometrics. Thus, the first group was integrated by 20 researchers in the mentioned areas from countries such as Chile, Argentina, Colombia, and Spain. On the other hand, the second group was ordinary people with no experience or knowledge in the research area. Therefore, the group of researchers allowed us to establish a control group for other participants' responses.

\begin{figure*}[!ht]
  \centering
  \includegraphics[width=0.98\textwidth]{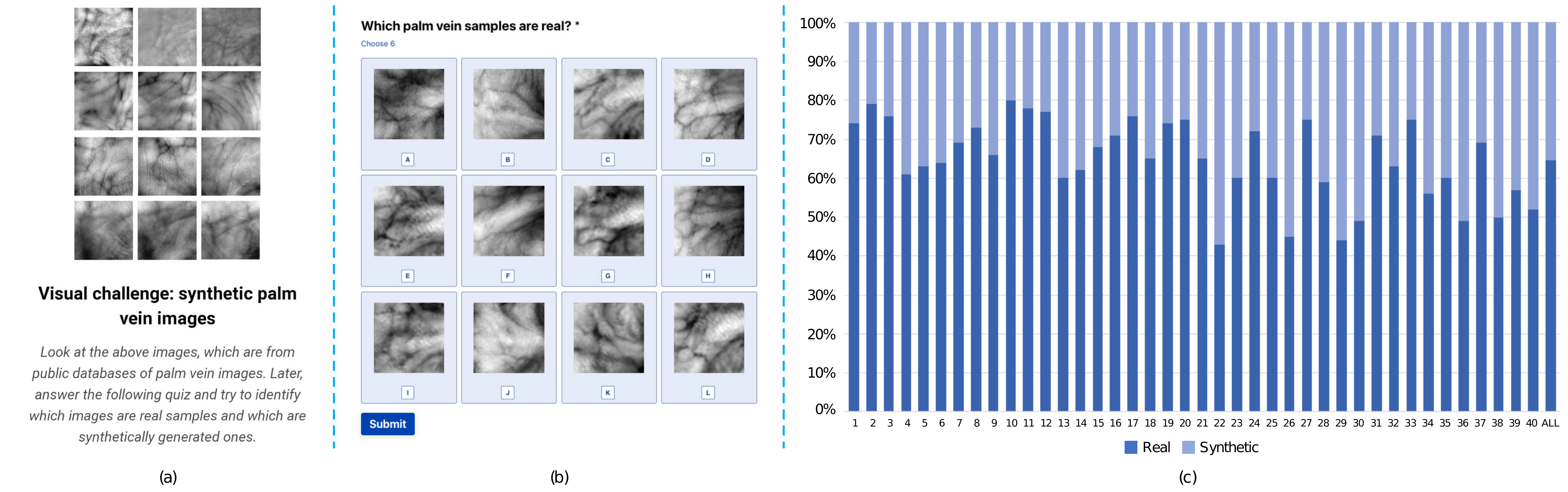}
  \caption{Examples of the visual inspection application (a)(b) and the corresponding bar chart from the participants' responses (c).}\label{f:visual_results}
\end{figure*}

To carry out the experiment, we created an online application to show palm vein samples to the participants and ask them to choose which from a gallery were real images. Initially, as depicted in Figure~\ref{f:visual_results}(a), the application showed a first screen displaying random samples from publicly available databases, where participants could familiarize themselves with the visual characteristics of this type of image. Later, a gallery of 12 random images was shown like in Figure~\ref{f:visual_results}(b), 6 of which correspond to real samples from public datasets, and the other 6 were from our synthetic databases. Then, each contestant had to choose six samples that he/she considered to be real.

Figure~\ref{f:visual_results}(c) shows a bar chart plotting the percentage of each type of sample that was classified as real by each participant. In the graph, numbers from 1 to 20 correspond to the control group of researchers, whereas from 21 to 40 are ordinary contestants, and the last bar is the responses' overall average. Table~\ref{t:visual_results} also summarizes the average percentages of each group. From the results, it is supposed that the higher the percentage of synthetic samples classified as real, the greater the similarity of these images. 

\begin{table}[ht!]
\centering
\caption{Average classification percentages of each group in the visual evaluation.}\label{t:visual_results}
\footnotesize
\begin{tabular}{ccc}
\toprule
\textbf{Group} & \textbf{Real Samples} & \textbf{Synthetic Samples} \\ \midrule
Researcher     & 70.55                 & 29.45                      \\ 
Ordinary       & 58.70                 & 41.30                      \\ 
All            & 64.63                 & 35.37                     \\
\bottomrule
\end{tabular}
\end{table}

\begin{figure*}[!ht]
  \centering
  \includegraphics[width=0.98\textwidth]{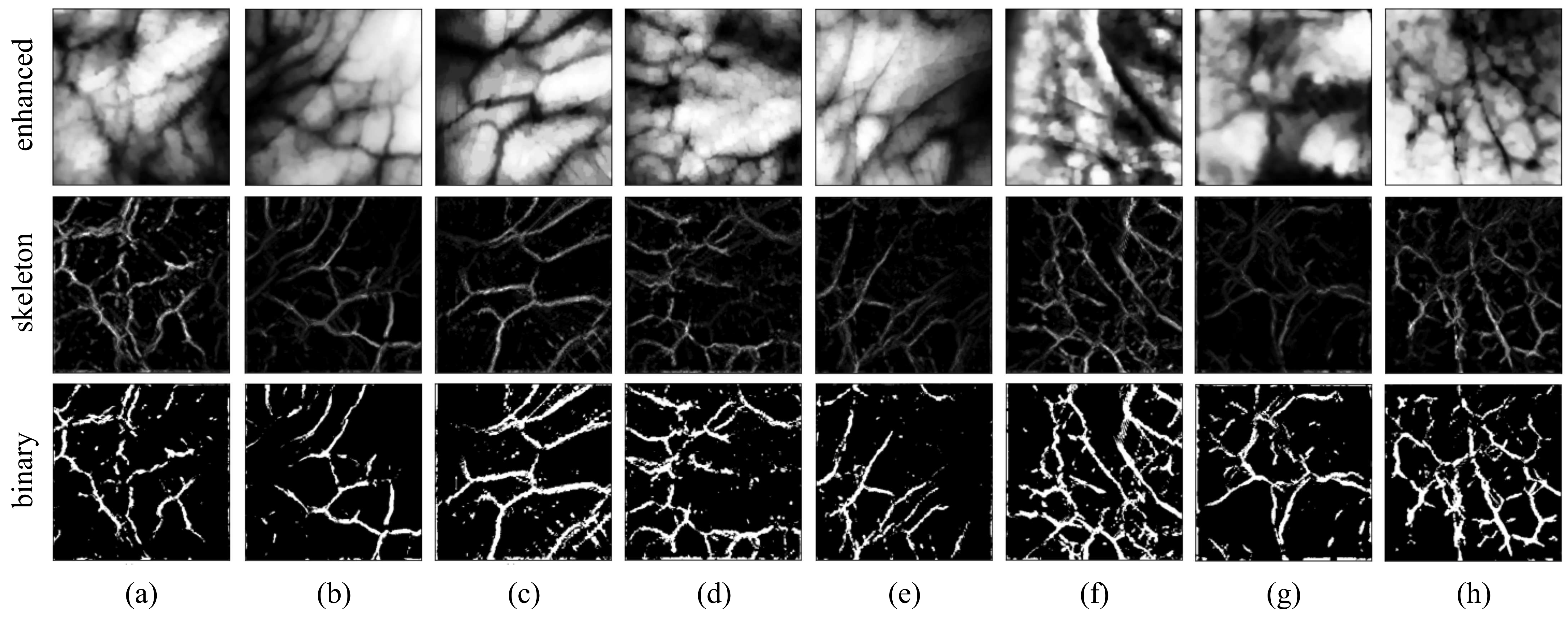}
  \caption{{ROI samples and the extracted vein patterns from different public datasets: (a) CASIA, (b) PUT, (c) PolyU, (d) Tongji, (e) IITI, (f) FYO, (g) Synthetic-sPVDB, and (h) NS-PVDB. The first row shows enhanced ROI images, while the second and third rows contain the skeletonization and binarization of vein patterns, respectively.}}\label{f:segmentation}
\end{figure*}

It can be noticed that although on average more than 40\% of the synthetic samples were classified as real by the ordinary group, this value was about 11\% lower for the control group. On the one hand, this behavior was partly influenced by the fact that the images were randomly displayed from the databases. Besides, according to previous qualitative and quantitative tests~\cite{icprs2021style,icprs2021nature}, the generated databases have more similarity to some of these. {In order to compare palm vein patterns from different public databases and the generated synthetic datasets, Figure~\ref{f:segmentation} shows ROI samples and the vein patterns extracted from them. It is possible to appreciate the existing visual similarities in the original and segmented images using skeletonization and binarization of the vein patterns, respectively. In addition, similar cyclic structures corresponding to the venous network of the hand are observed in the vein patterns obtained.} Although the two types of images have certain visual similarities, it is worth noting that unrealistic details can be presented in some cases. Consequently, a small grid effect can be observed for the images produced by the Style-VIG generator. In contrast, non-homogeneous textures of vein patterns are produced in the samples generated by the Nature-VIG procedure. 

\subsubsection{Similarity Analysis based on Probability Densities of Pixels Distribution}

In the literature, different works in the area of synthetic biometric imaging~\cite{zuo2007generation,hillerstrom} use the Bessel K-forms~\cite{srivastava2002stochastic} as a method to compare the lower-order similarity of images textures. Thus, this section establishes a texture-level distinction between real and synthetic samples following the same methodology. The method models the lower-order probability densities of bandpass filtered images and allows comparison of the studied images~\cite{zuo2007generation}. The Bessel k-forms is a parametric family that depends on two parameters: shape $p$ and scale $c$, as shown in Equation~\ref{eq:kforms}:

\begin{equation}\label{eq:kforms}
f(x;p,c)=\frac{1}{Z(p,c)}|x|^{p-0.5}K_{p-0.5}\left(\sqrt{\frac{2}{c}}|x|\right),
\end{equation}
where, $K_{(p-0.5)}(\sqrt{2/c}|x|)$ denotes modified Bessel functions of the second kind and $Z$ is the normalizing constant given by $Z(p,c)=\sqrt{\pi}\Gamma(p)(2c)^{0.5p+0.25}$ ~\cite{srivastava2002stochastic}. 

It is worth mentioning that each filter selects and isolates certain features present in the original image, and they depend on the specific task. For K-form modeling, in~\cite{srivastava2002universal} suggest filters whose resulting spectral components have marginals with the following characteristics: 1) unimodal with the mode at zero, 2) symmetric around zero, and 3) leptokurtic. Particularly, in this analysis, we use the log-Gabor filter, as well as in~\cite{hillerstrom}. The estimation of the parameters $p$ and $c$ for the K-forms can be determined from the observed data by the expressions:

\begin{equation}\label{eq:bc}
\hat{p}=\frac{3}{SK(I)-3}\ \ \ \text{and}\ \ \ \hat{c}=\frac{SV(I)}{\hat{p}},
\end{equation}
where, $SK$ and $SV$ are the sample kurtosis and variance of the filtered image $I$, respectively. 

\begin{figure*}[!ht]
  \centering
  \includegraphics[width=0.8\textwidth]{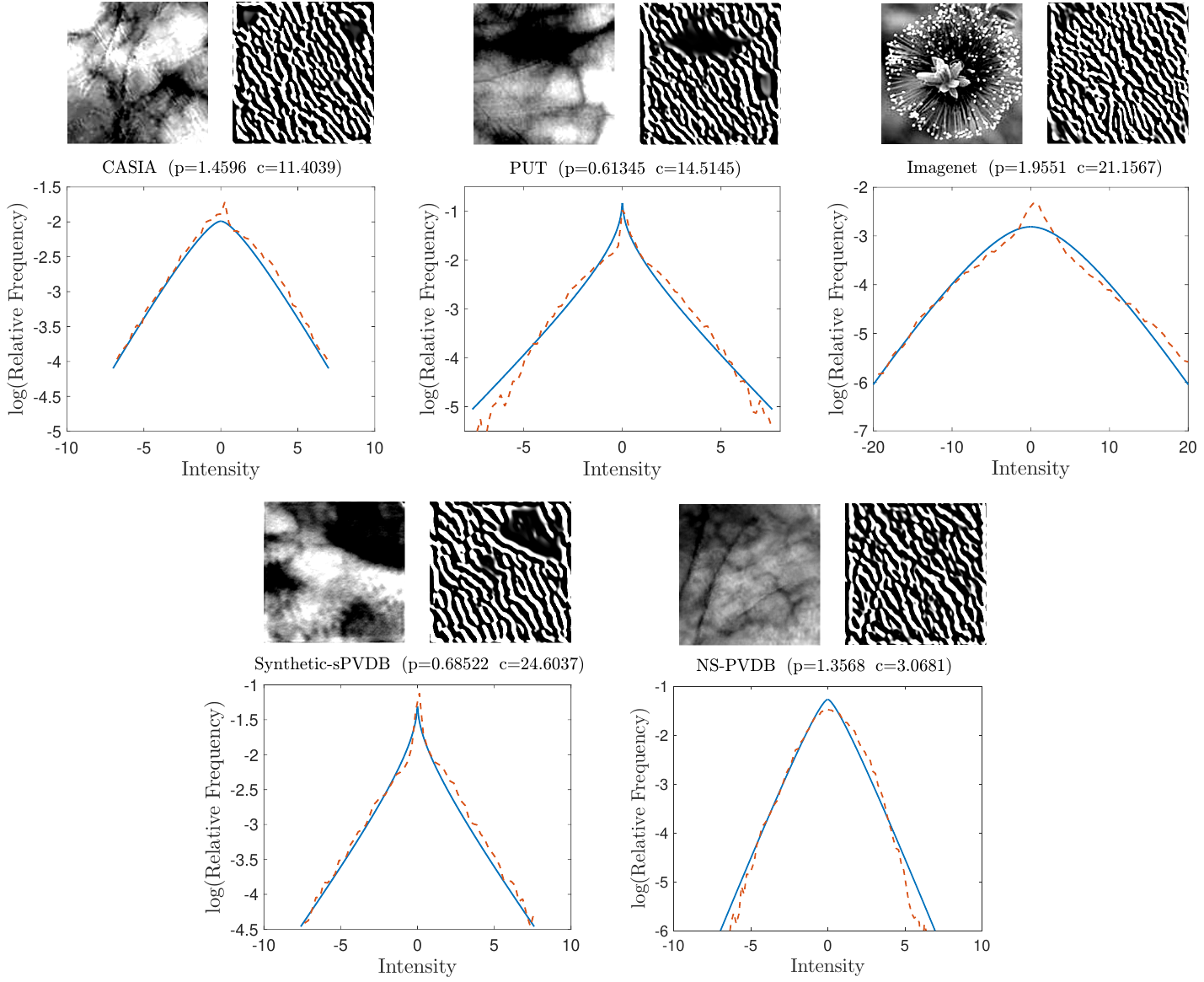}
  \caption{Comparison between real (first row) and synthetic (second row) samples using the Bessel K-forms. Each sub-image shows the original sample, the filtered image, and the corresponding curve of pixel density. Continuous and discrete trace curves represent the estimated K-forms and observed density functions from each image, respectively.}\label{f:kforms}
\end{figure*}

To compare real and generated palm vein images, we use the Bessel K-shape representation, as it allows us to quantify the degree of disorder of the veins. For the initial comparative analysis, we randomly selected ten samples from five different datasets: synthetic palm vein samples from the Synthetic-sPVDB and NS-PVDB databases, real images from the PUT~\cite{PUT} and CASIA~\cite{CASIA} datasets, and natural images from the Imagenet database~\cite{deng2009imagenet}. The images were scaled to the same size, and the intensities of gray levels were normalized, and then the log-Gabor filter was applied. Figure~\ref{f:kforms} comparatively show examples of each set of images with their corresponding filtered image and the associated curves with the observed and estimated pixel densities with the Bessel K-forms.

Table~\ref{t:kforms} presents the values of the parameters of Bessel K-forms estimated from the filtered images.  It can be seen that the shape parameter $p$ oscillates in a wide range from a synthetic image to a real image, but this same behavior can be observed in the CASIA database, whose values range from 0.38 to 2.38. This behavior is reasonable since the palm vein patterns do not have a particular shape, and also, the images are influenced by many parameters as explained in Section~\ref{ss:acquisition}. As for the value of the $c$ parameter, in~\cite{srivastava2002universal} they show that it is not relevant since it essentially represents a scale parameter related to the range of pixel values in the image.

\begin{table}[ht!]
\centering
\caption{Estimated Bessel K parameters for different images from the evaluated databases. The last row corresponds to the average values and the corresponding standard deviation.}\label{t:kforms}
\scalebox{0.73}{
\footnotesize
\begin{tabular}{cccccccccc}
\\ \hline
\multicolumn{4}{c}{\textbf{Synthetic}} & \multicolumn{4}{c}{\textbf{Real}}  &  \multicolumn{2}{c}{\textbf{Natural}}  \\ \midrule
\multicolumn{2}{c}{\textbf{sPVDB}} & \multicolumn{2}{c}{\textbf{NS-PVDB}} & \multicolumn{2}{c}{\textbf{PUT}}  & \multicolumn{2}{c}{\textbf{CASIA}}  &  \multicolumn{2}{c}{\textbf{Imagenet}}  \\ \midrule
$p$ & $c$ & $p$ & $c$ & $p$ & $c$ & $p$ & $c$&$p$&$c$\\ \midrule
0.6849 & 24.5930 & 1.1487 & 10.1912 & 0.6134 & 14.5145 & 1.4596 & 11.4039 & 0.8477 & 88.2131 \\ 
0.5273 & 28.6493 & 1.3568 & 3.0681 &0.4633 & 21.5892 &  0.5022 & 18.2254 & 2.5483 & 24.3118 \\ 
1.1516 & 12.7055 & 1.2071 &  5.3395 & 0.4919 & 20.7747 &  0.3798 & 23.8390 & 2.9419 & 11.3912\\ 
0.6418 & 15.9923 & 1.3568 &  3.0681 & 0.3637 & 29.3623 & 0.3906 & 5.0458 & 1.8931 & 21.9251 \\ 
0.2960 & 40.8959 & 1.3655 & 7.0201 & 0.3495 & 32.8473 & 1.4355  & 7.7786 & 2.5702 & 43.5869\\ 
0.9113 & 10.9084 & 0.7042 & 13.0153 & 0.3112 & 44.0356 & 0.6174 & 23.4266& 0.6247 & 180.5680  \\ 
0.8797 & 10.9918 & 0.7032 & 12.2776 & 0.5808 & 19.0018 & 0.7953 & 12.1579 & 0.6241 &  126.4837 \\ 
0.5773 & 35.3900 & 0.8982 &  7.2657 & 0.3233 & 35.5480 & 0.5964 & 20.8183 & 2.7680 & 31.9808  \\ 
0.3800 & 36.2374 & 1.0111 & 7.9521 & 0.2980 & 40.1542 & 0.4522  & 13.5832 & 0.9053 & 64.5027  \\ 
0.6298 & 15.8142 & 0.9391 &  7.6821 & 0.3852 & 34.7648 & 1.3945 & 5.1034 & 2.0420 & 56.7189 \\ \hline
0.67 ($\pm$ 0.26) & 23.22 ($\pm$ 11.45) & 1.07 ($\pm$ 0.26) & 7.69 ($\pm$ 3.41) & 0.42 ($\pm$ 0.11) & 29.26 ($\pm$ 9.85) & 0.80 ($\pm$ 0.45) & 14.14 ($\pm$ 7.13) & 1.77 ($\pm$ 0.94) & 64.97 ($\pm$ 53.41) \\
\bottomrule
\end{tabular}
}
\end{table}

Due to the above, using the $p$ and $c$ parameters is not indicative of a meaningful comparison between real and synthetic palm vein images. Hence, to quantify the difference between pairs of filtered images based on the Bessel K-forms, we explored two distance measures, the $KL$ divergence and the $L^2$-metric defined as~\cite{srivastava2002stochastic}:

\begin{eqnarray}
d_{KL}&=&\int_{IR}\log\left(\frac{f_1(x)}{f_2(x)}\right)f_1(x)dx\label{eq:dKL}\\
d_I&=&\sqrt{\int_x(f_1(x)- f_2(x))^2}dx,\label{eq:dI}
\end{eqnarray}
where, $f_1(x)$ and $f_2(x)$ are the functions corresponding to the K-forms of each image. 

\begin{figure*}[!ht]
  \centering
  \includegraphics[width=0.45\textwidth]{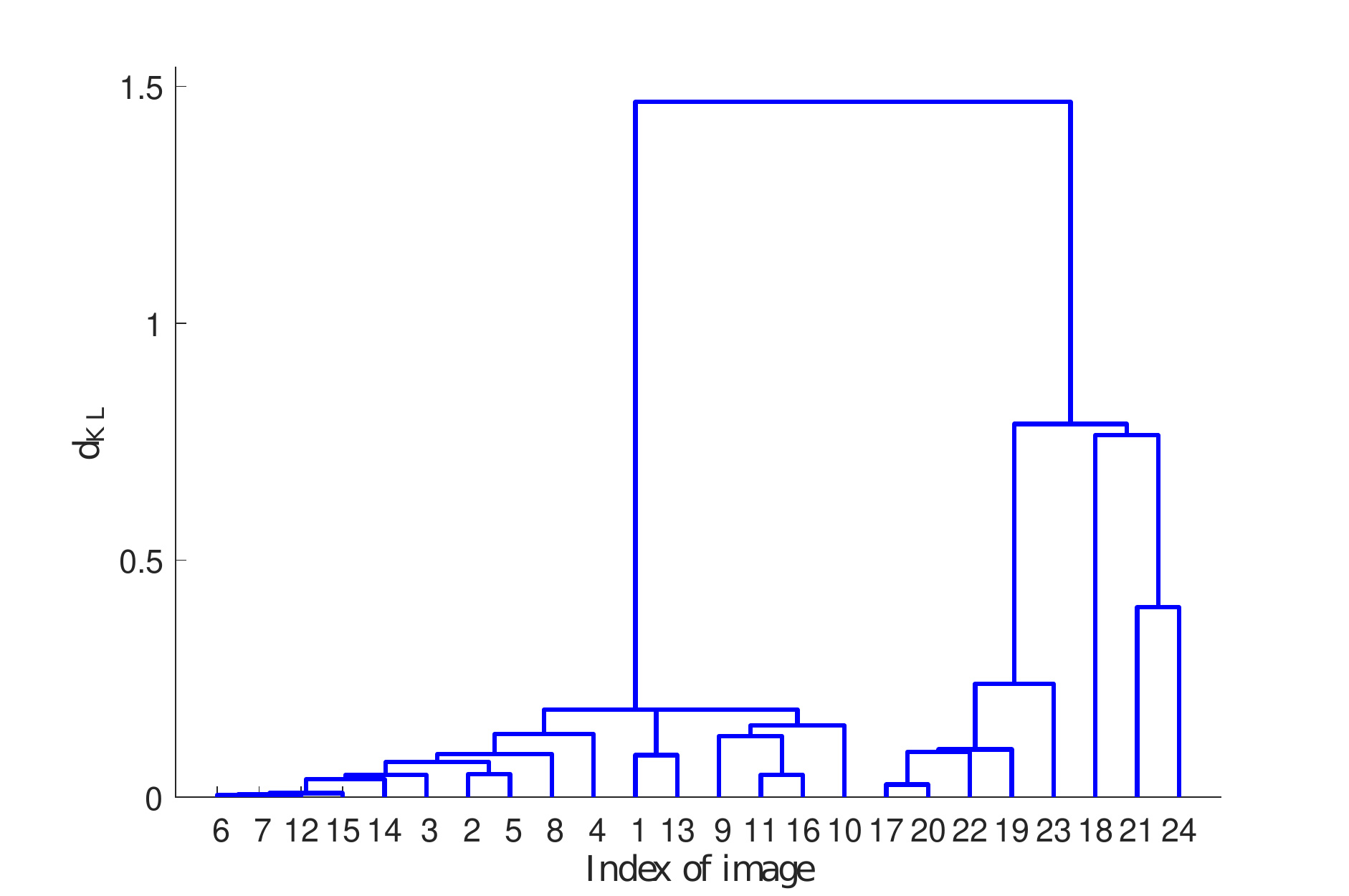}
  \includegraphics[width=0.45\textwidth]{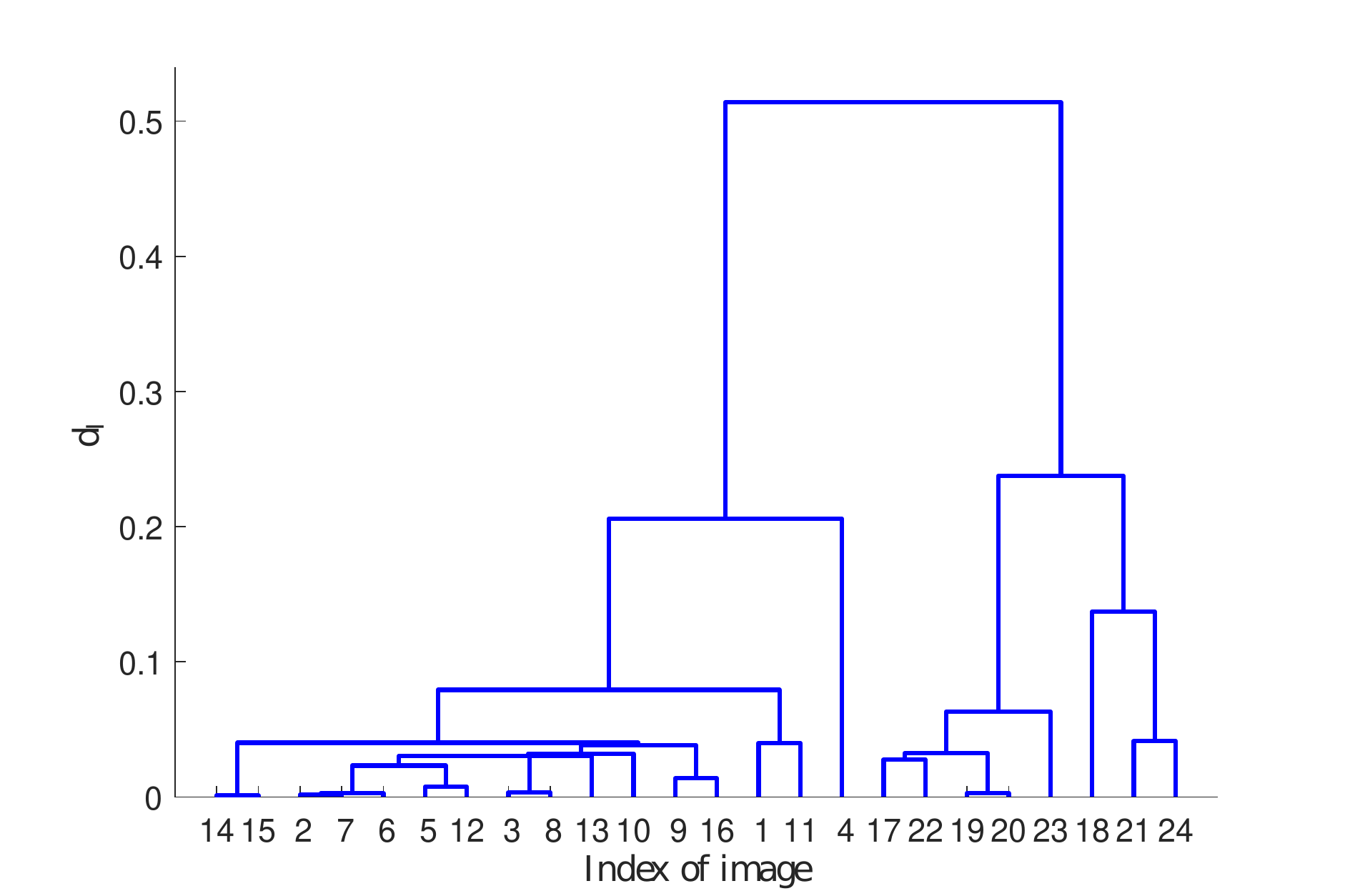}
  \caption{Clustering dendrogram plot of the distance measures $d_{KL}$ (left) and $d_{I}$ (right) obtained from the classification of images of public datasets (1-8), Synthetic-sPVDB database (9-16), and natural images of Imagenet database (17-24).}\label{f:style850put}
\end{figure*}

\begin{figure*}[!ht]
  \centering
  \includegraphics[width=0.45\textwidth]{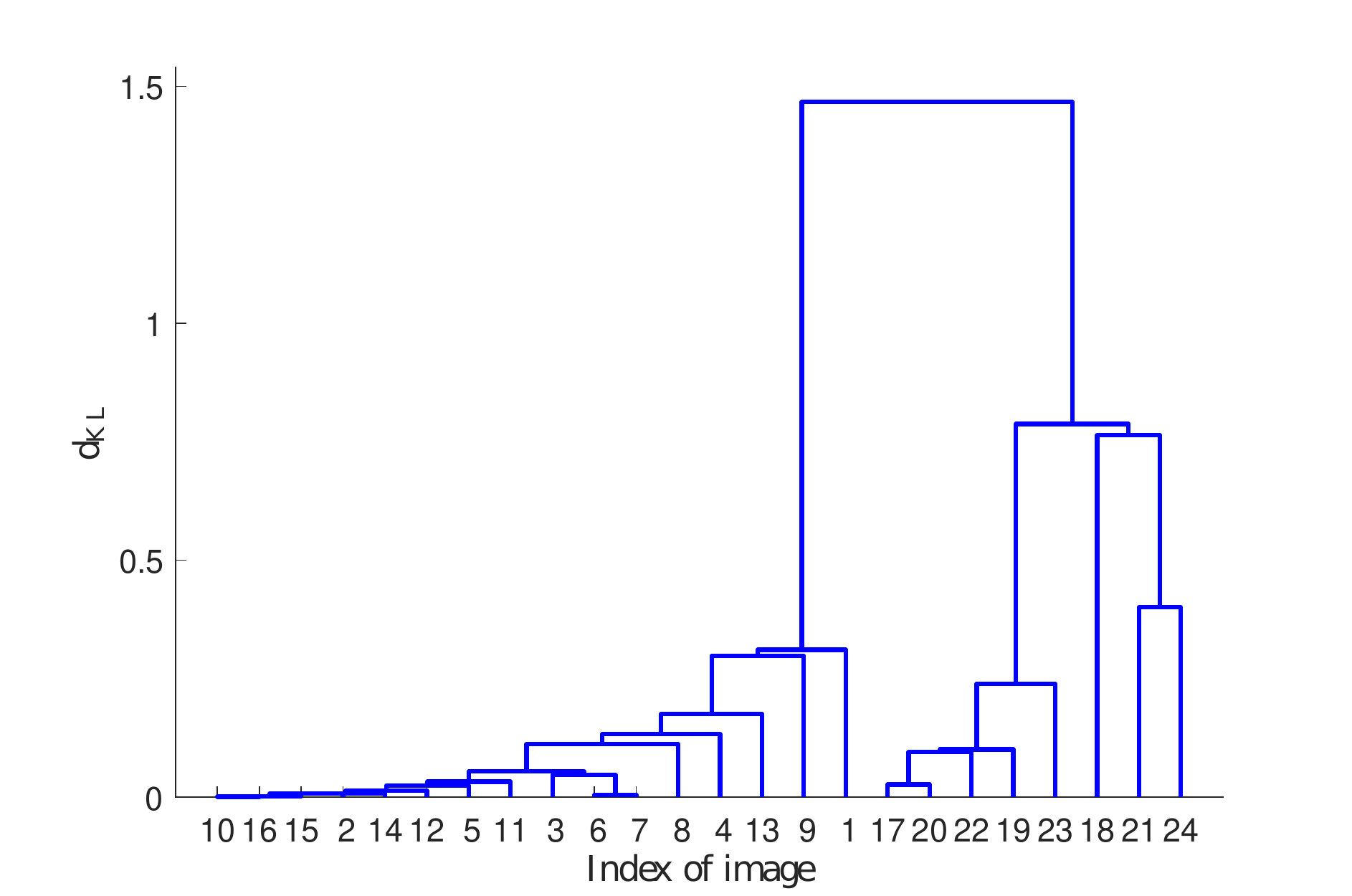}
  \includegraphics[width=0.45\textwidth]{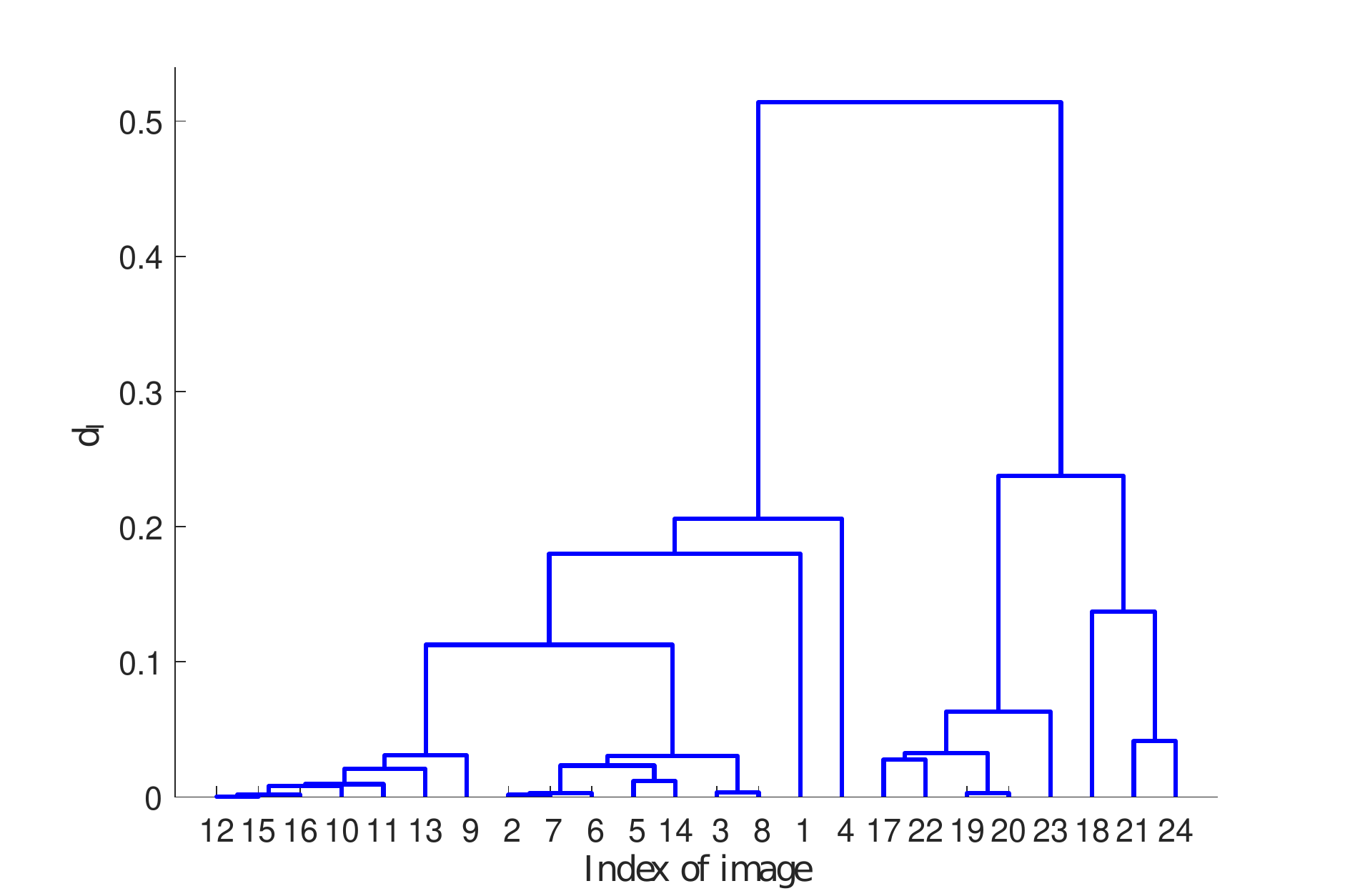}
  \caption{Clustering dendrogram plot of the distance measures $d_{KL}$ (left) and $d_{I}$ (right) obtained from the classification of images of public datasets (1-8), NS-PVDB database (9-16), and natural images of Imagenet database (17-24).}\label{f:physarum850put}
\end{figure*}

The distances defined above allow to compare and classify the Bessel K-forms of the three groups of evaluated images: synthetic, real, and natural, as well as in~\cite{hillerstrom}. Thus, the distance matrices among different types of images were calculated, and we performed a hierarchical clustering. Figures~\ref{f:style850put} and~\ref{f:physarum850put} depict the resulting dendrogram plots obtained from the classification of images of the Synthetic-sPVDB and NS-PVDB databases, respectively. In the plots, indices 1-8 denote real images taken from CASIA and PUT databases, 9-16 represent synthetic images, and 17-24 correspond to natural images. It can be noticed that real and synthetic images are clearly distinguished from natural images in all cases. Moreover, real and synthetic images are grouped into similar branches.

\subsubsection{Quantitative evaluation}

In this section, we quantitatively evaluate the proposed synthetic databases against the most representative public datasets of the state-of-the-art. For this purpose, we used quantitative measures to estimate the differences in the feature space, which is more useful than the pixel space for biometric tasks. Since the databases have different sizes, we randomly sample the 10\% of images from each of them for comparison. The images were represented in the feature space using the final activations of an Inception model~\cite{InceptionV3}. 

First, we computed the FID metric~\cite{FID} to measure the statistical difference between the distributions of feature vectors. The image features are modeled using multidimensional Gaussian distributions $\mathcal{N}(\mu,\Sigma)$ with mean $\mu$ and covariance $\Sigma$. Thus, let $x$ be a set of real images and $g$ be a set of generated images, the $FID(x,g)$ is computed as follows:

\begin{equation}\label{eq:FID}
    FID(x,g)=\left \| \mu_x-\mu_g \right \|^2_2+\operatorname{Tr}(\Sigma_x+\Sigma_g-2(\Sigma_x\Sigma_g)^{\frac{1}{2}}),
\end{equation}
where $\operatorname{Tr}$ is trace operator defined as the sum of all the diagonal elements. In the case of our proposed databases, the FID metric is fundamental since it is a standard for evaluating the quality of images generated by GAN models. The lower the FID value, the better the quality and diversity of the generated images. 

Besides, we computationally replicate the visual inspection experiment described in Section~\ref{ss:visual_results} implementing a binary classifier. Therefore, we used a 1-Nearest Neighbour (NN) classifier with a Leave-One-Out (LOO) scheme as a simple method to classify real and synthetic samples. In this case, it is expected that the closer the Accuracy is to 50\%, the greater the similarity of the synthetic images are against real ones. In addition to the classification accuracy, we applied the PRD method proposed in~\cite{PRD} to compute the precision ($F_{1/8}$) and recall ($F_8$) metrics based on the feature distributions. These metrics allow the evaluation of GAN-based models complementing the FID metric. The main idea is to quantify how much of the real distribution can be generated by a part of the synthetic distribution and vice-versa.

Table~\ref{t:metrics} shows the results of quantitative evaluation of both generated databases against the most representative state-of-the-art public datasets. The obtained results suggest that the synthetic images perform better for specific databases depending on the generation procedure. Thus, Synthetic-sPVDB is closely related to CASIA-940 due to the Style-VIG process was trained with images from CASIA-850. On the other hand, NS-PVDB has more similarities to PUT and VERA datasets because their images visualize the vein patterns more clearly. These results are also consistent with the visual inspection performed.

\begin{table}[ht!]
\centering
\caption{Results of quantitative evaluation of both generated databases (a) Synthetic-sPVDB and (b) NS-PVDB, against the most representative public datasets of the state-of-the-art.}\label{t:metrics}
\footnotesize
\begin{tabular}{lcccccccc}
\toprule
\multirow{2}{*}{\textbf{Real Dataset}} & \multicolumn{4}{c}{\textbf{Synthetic-sPVDB}} & \multicolumn{4}{c}{\textbf{NS-PVDB}}  \\ \cline{2-9}
 & \textbf{FID} & \textbf{Accuracy} & $\mathbf{F_8}$ & $\mathbf{F_{1/8}}$ & \textbf{FID} & \textbf{Accuracy} & $\mathbf{F_8}$ & $\mathbf{F_{1/8}}$  \\ \midrule
CASIA-940~\cite{CASIA} & 67.54 & 0.604 & 0.897 & 0.898 & 227.41 & 0.841 & 0.295 & 0.318 \\
PolyU~\cite{PolyU} & 103.60 & 0.665 & 0.743 & 0.834 & 203.60 & 0.815 & 0.348 & 0.322 \\ 
PUT~\cite{PUT} & 209.11 & 0.896 & 0.368 & 0.309 & 71.92 & 0.719 & 0.641 & 0.737 \\ 
VERA~\cite{VERA} & 93.01 & 0.637 & 0.750 & 0.844 & 93.01 & 0.634 & 0.825 & 0.865 \\ 
Tongji~\cite{Tongji} & 88.91 & 0.621 & 0.853 & 0.862 & 153.01 & 0.714 & 0.403 & 0.447 \\ 
IITI~\cite{IITIpalmvein} & 97.18 & 0.664 & 0.727 & 0.814 & 137.01 & 0.672 & 0.371 & 0.386 \\ \bottomrule
\end{tabular}
\end{table}

\section{Discussion, Challenges, and Future Research Directions}\label{s:discussion}

\subsection{Discussion and Main Challenges}

The identification of people by palm veins provides a high level of security since the vascular structure is not exposed to the environment, and the hands generally remain in a half-open state. Compared to the veins of the fingers or the veins of the back of the hand, the palm has complex vascular patterns that generate many differentiating characteristics for personal identification. On the other hand, within the current context of Covid-19, the contactless devices that this biometry uses have been a hygienic and non-invasive solution that promotes a high level of acceptance by the user.

The main disadvantage of palm vein biometry is associated with the reduced number of subjects in public databases because they avoid the evaluation of the reliability and scalability of biometric recognition algorithms~\cite{wu2019}. Therefore, in this paper, we review the parameters related to the acquisition devices, soft biometrics, and the anatomy of the palm veins to deepen the generation of synthetic images of the veins of the palm. From this perspective, the construction of a large-scale palm vein database using synthetic imaging mainly faces three challenges, which are described below:

\begin{itemize}

    \item[(i)] Vascular structure: the vascular networks of the hand have complex patterns to ensure blood flow in the event of any movement. However, the multiple factors involved in obtaining images allow only the capturing of discontinuous patterns that avoid the complete segmentation of the network and its respective structural analysis. There are imaging techniques such as magnetic resonance angiography or hyperspectral images to visualize vein networks with greater precision. Despite that, these techniques are expensive and therefore not feasible to produce a biometric database and establish structural relationships. Consequently, to reproduce the vascular structure of the palm, theoretical knowledge associated with the anatomy of the arterial network and the venous network is required. The arterial part has a branched structure described in detail, but the venous part is only described as a random network that obeys optimization principles.

    \item[(ii)] Palm texture: to reproduce the palm texture, factors associated with the acquisition devices that produce variation in brightness and contrast must be considered. Several aspects influence this variability: soft biometrics, local contrast, and lighting conditions. Soft biometrics such as gender, age, body build, and ethnicity impact in the NIR lighting response \cite{Xie2018palmprint, Damak.etal2019palm, zheng2017static, edelman2012identification}. Among the public databases, only the VERA database collects extra information about the sex and age of the individuals, which limits the study of these factors on the variability of the image. On the other hand, local contrast occurs for different reasons; among these are the different aesthetic units of the palm, the marks or defects on the skin, and the hand's position at the time of image capture. Finally, lighting conditions such as the NIR spectrum, environmental factors, body temperature, and light intensity directly affect the acquisition of the palm vein pattern since they generally produce low contrast, high noise level, and uneven lighting. It should be noted that both the variation of the power and the variation of the NIR spectrum from one acquisition device to another produce a non-standardized technology and the generation of images with different qualities.

    \item[(iii)] Quality of palm vein images: a synthetic database of palm veins should guarantee that the intra-class and inter-class variabilities are in a quality range that allows the evaluation of the reliability and robustness of the biometric recognition algorithms. In real images of palm veins, the evaluation of the quality ranges has been poorly studied, and consequently, there is not yet a standard that defines them. Regarding the works related to quality evaluation, these are limited to the analysis of self-created images and not the available public databases. Therefore, it avoids establishing a quality range that allows the scientific community to compare the robustness of the developed systems.

\end{itemize}

In Section~\ref{ss:RedesBiometria} the existing models that allow the generation of synthetic images for biometric purposes were described. According to the review considered in this article, the models can be classified into texture methods and structural methods. The texture methods consider probability functions to reproduce the different gray levels of the images. These models are used in the augmentation of samples and not in the generation of new images, which tends to work with verification algorithms rather than identification algorithms. In contrast to the above, structural models generate new images by reproducing properties of the vascular network and the effects of acquisition devices. The importance of structural models is that they allow the evaluation of identification algorithms. Regarding the existing structural models, only the works developed in \cite{crisan2008, hillerstrom} simulate vein images of the back of the hand and fingers, sharing similarities with palm vein images. However, the palm has more details at the level of vascular structure and palm texture, which obstructs their respective simulation.

Given the lack of knowledge of the palm vein structure, in Section~\ref{ss:nature}, the different models that allow the generation of biological networks were reviewed in order to establish analogies with the structures found in the palm. Two types of structures are distinguished in biological networks: 1)~open structures, responsible for the main nutrient supply, and 2)~closed structures, responsible for fault-tolerant flow. Although there are different methods to generate these structures, they have been mainly inspired by the growth of tumors, the development of the leaves veins, or the growth of some unicellular organisms. Among the applications of these models are retinal veins, venation of insect wings, and topological optimization for heat flow; however, none of the existing models have been used to simulate palm vein structures. Despite that, the existing models could be adapted by means of appropriate parameters that make it possible to simulate the characteristics of the venous network of the palm.

In Table~\ref{P:palmvein}, we present a classification of existing models depending on the generation of vascular networks based on the arterial and venous structure of the palm. Models based on reaction-diffusion and diffusion-limited aggregation only allow the generation of venous structures since the bifurcations of the network cannot be controlled because they arise from random processes. The CCO algorithm would only allow generating the superficial arterial structures of the palm because the network is obtained by adding segments guided by an optimization objective that depends on the pressure or flow at the network terminations, which avoids the formation of loops that are characteristic of the venous network. The remaining models can generate both types of structure. However, they have disadvantages in reproducing palm vein patterns, except pathfinding with adaptive dynamics search, spatial colonization algorithm, and hybrids methods,  because the parameters used in the growth of the network can be redefined to generate a venous pattern with an appropriate shape and besides consider optimality principles. On the other hand, the active shape needs good quality images to reproduce each network. The \emph{Lindenmayer system} and \emph{Particle system} can produce random structures associated with the venous network, but they do not take optimization principles into account. Although it was clear that the \emph{Lindenmayer systems} also generate fractal structures that can be associated with genetic development, they do not allow the modeling of environmental conditions.  Therefore, the most suitable models are pathfinding with adaptive dynamics search, spatial colonization algorithm, and hybrids methods, as they allow for the generation of both types of vascular networks.  Finally, the B-spline-based vascular models are a technique that allows efficient point joining with the characteristic that the resulting curve is differentiable at all points, thus providing images with a high degree of realism.  Thus, it could be a good combination with the previous models.

\begin{table}[ht!]
  \centering
  \caption{Classification of existing models depending on the generation of vascular networks of the hand.}\label{P:palmvein}
  \footnotesize
  \begin{tabular}{lcc}
    \\ \toprule
    \textbf{Model} & \textbf{Arterial structure (known)} & \textbf{Venous structure (random)} \\  \midrule
    Reaction-diffusion~\cite{turing} & No & Yes  \\
    Pathfinding with adaptive dynamics~\cite{jones2010characteristics} & Yes & Yes, (Optimal shape) \\
    Constrained Constructive Optimization~\cite{de2015} & Yes & No  \\
    Diffusion-limited aggregation~\cite{witten1981} & No & Approximate  \\
    Active shape~\cite{cootes1995active} & Yes & Yes  \\
    Particle system~\cite{rodkaew2002algorithm} & Yes & Approximate \\
    Lindenmayer system~\cite{prusinkiewicz} & Yes, radial part (fractal shape) & Yes, fractal shape \\
    Space colonization algorithm~\cite{runions2007} & Yes & Yes  \\
    Hybrid~\cite{jin2009improved,salcedo2019hybrid} & Yes & Yes  \\
    B-spline based vascular structures~\cite{li2007mathematical,castro2020visual} & Yes & Yes \\
    \bottomrule
  \end{tabular}
\end{table}


Based on previous works and the review of the state-of-the-art, in Section~\ref{sec:results} we introduced a general flowchart for the creation of a synthetic palm vein database. The presented model has three key foundations: (1) the vein image generator process can be implemented using different approaches depending on the parameters involved; (2) the scheme considers a uniqueness discriminator based on similarity matching, which allows the continuous growth of the biometric database; and (3) several samples per each individual comprise the generated database by applying random transformations to simulate natural variations of the acquisition process. Additionally, two generalist approaches for creating synthetic palm vein images were examined. The conducted analysis provides valuable support for the present review and strengthens the knowledge foundation for future research in this area.

The first approach consists of generating palm vein images using the StyleGAN2 architecture. Data augmentation was performed before training the network to prevent any generated sample from being rejected by the discriminator due to overfitting. However, the generator would also learn to produce the augmented data distribution that may not correspond to the real images. Hence, other generative models such as different GAN architectures or the VQ-VAE model should be evaluated to code more complex characteristics related to palm vein patterns. In the second approach, texture and vascular structure are independently generated and then fused. The vascular network is generated through a nature-based optimization process, which may not correspond to the real palm vein structure. Although both approaches validate the idea for the generation of synthetic palm images, they do not consider specific characteristics of the palm vascular structure. Therefore, they only represent the initial milestone towards developing more complex methods, which allow the generation of vein images considering the multiple parameters influencing the visualization of vein patterns. Furthermore, the computation time increases as the data increases due to the individual discriminator when constructing a large-scale database. Thus, parallel and distributed computing must be considered to implement the IUD process.

{Finally, in Section~\ref{ss:acquisition}, we analyze that palm vein acquisition devices are sensitive to soft traits of people, so it is necessary to consider them in the generation of synthetic images. Incorporating soft biometrics would benefit image realism and improve the evaluation performance of recognition algorithms. However, to the best of our knowledge, studies associated with the influence of soft biometrics on vein pattern visualization are very preliminary, and these studies do not allow for accurate parameter estimation for their respective simulation. Going deeper into the influence of soft biometrics requires databases containing this type of information. Therefore, further and more in-depth studies could be conducted to estimate appropriate parameters and consider them in the generation model through certain state variables that evaluate the variability of the visualization of vein patterns.}

\subsection{Future Research Directions}

According to the problems faced by vascular-based biometrics and the generation of synthetic palm vein images, from our humble perspective, the following lines of investigation can be considered by the research community:

\begin{itemize}

   \item[(i)] Knowledge of the vascular network: Considering that the structure of the palm vein network is not known in detail and that the images used in biometric recognition only provide discontinuous patterns of the vascular network, it is necessary to deepen its characterization to establish parameters that allow the network modeling. This characterization would also allow the total reconstruction of the network in the images, which would improve, among other things, the performance of the biometric recognition algorithms.

   \item[(ii)] Construction of databases that incorporate soft biometrics: The development of databases that collect extra information such as gender, age, medical conditions, etc., would allow a detailed study of its influence on the visualization of palm veins. These studies could improve the quality of the images since there would be criteria for adjusting the lighting parameters of the acquisition devices. 
   
   In addition, this extra information from the images of the palm veins will allow the creation of soft biometric recognition models and multitasking biometric recognition, which could be used for other types of applications, such as for non-invasive health studies.
   
   \item[(iii)] Image quality evaluation: Regarding the problem associated with the lack of a quality evaluation standard for palm vein images, it is required to analyze the quality of images from public databases and define the range of variability both intra-class and inter-class to provide a reference framework available to the scientific community.
   
   \item[(iv)] Cross-spectral matching: due to the non-standardization of palm vein acquisition methods, there is a need to propose robust algorithms at different wavelengths of NIR illumination to recognize people regardless of the NIR spectral band used in the captured image.
   
   \item[(v)] Modeling synthetic images of palm veins: It is necessary to consider factors that affect the representation of vein patterns to guarantee the realism of the images generated. First, an algorithm must be proposed that reproduces the vascular structure of the palm, for which we could consider variants of the existing vascular network models studied in Section~\ref{s:review}. On the other hand, it is required to reproduce the texture of the palm, which could be done using generative models, taking into account the effects generated by the acquisition devices and the variability of the quality of the images.
   
   \item[(vi)] Creation of large-scale databases: With the problems associated with the limited number of individuals in public databases, the simulation of the vascular network of the palm based on the anatomy and the simulation of the effects produced by the acquisition systems would allow developing a database of synthetic images using the proposed scheme in Figure~\ref{f:model}. It would allow the evaluation of palm vein biometric systems on a large scale.
   
\end{itemize}

\section{Conclusions}\label{s:conclusions}

The objective of our study was to address the problems associated with the limited number of images in public databases, which prevent the evaluation of the reliability and scalability of large-scale palm vein recognition algorithms. To address this problem, the present work analyzed different approaches of the state-of-the-art for the generation of vascular structures, allowing to establish the main limitations and advantages of the methods studied for the generation of the vascular network of the palm. The study was carried out from a mixed approach that analyzed both the factors that influence the representation of real palm vein images and the modeling of synthetic images of vascular structures.

Regarding the representation of the images of vein patterns, the anatomy of the palm veins was taken into account to determine geometric and topological characteristics that allow the generation of realistic vascular structures. In addition, we studied the factors that influence capturing the palm vein images and their quality evaluation to guarantee that the intra-class and inter-class variability are equivalent to the real images. Regarding the modeling of vascular structures, we reviewed the models reported in the literature to determine their flexibility to produce different structures that could be used in the generation of the vascular networks of the palm. Furthermore, we formalized a general scheme for the creation of a synthetic palm vein database.

Our study made it possible to identify the main challenges facing the generation of synthetic images of palm veins. By studying the models, it was possible to establish their advantages for the generation of the arterial and venous structure of the palm for biometric purposes. The evaluation of the proposed scheme for generating a large-scale synthetic database allowed validating the proposal's effectiveness and creating the largest palm vein databases of the state-of-the-art. Based on our findings, possible lines of research around the studied problem were proposed. Due to the limits of our perspectives, the suggestions we mentioned about future research are only for reference so that they could be enriched considering the comments and experience of the scientific community.

\section*{Acknowledgments}
This work was funded by the National Agency for Research and Development (ANID), Scholarship Program, Doctorado Becas Chile, 2022-21220765. R.H.-G. thanks to the Research Project ANID FONDECYT Iniciación en Investigación 2022 No. 11220693 “End-to-end multi-task learning framework for individuals identification through palm vein patterns”. 
The authors of the paper also thank to the Research Project FONDECYT REGULAR 2020 No. 1200810 “Very Large Fingerprint Classification based on a Fast and Distributed Extreme Learning Machine”. 
Portions of the research in this paper used the CASIA-MS-PalmprintV1 collected by the Chinese Academy of Sciences' Institute of Automation (CASIA). Portions of the research in this paper used the VERA-Palmvein Corpus made available by the Idiap Research Institute, Martigny, Switzerland.

\bibliographystyle{unsrt} 
\bibliography{references}

\end{document}